\PassOptionsToPackage{table}{xcolor}
\documentclass{template}
\newcommand{\templateoption}{option3}
\usepackage[T1]{fontenc}
\usepackage{lmodern}
\usepackage{xspace}
\usepackage{booktabs}
\usepackage{makecell}
\usepackage{graphicx,amsmath,amssymb,hyperref}
\usepackage[authoryear,round]{natbib}
\usepackage{xcolor}
\usepackage{wrapfig}
\usepackage[percent]{overpic}
\usepackage{url}
\usepackage{verbatim}

\hypersetup{
  colorlinks=true,
  urlcolor=magenta,
  linkcolor=black,
  citecolor=link,
  filecolor=black,
  pdfborder={0 0 0}
}
\definecolor{link}{HTML}{0063BE}
\providecommand{\equationname}{Equation}
\providecommand{\sectionname}{Section}
\providecommand{\appendixname}{Appendix}

\newcommand{\figref}[1]{%
  \figurename~\hyperref[#1]{\textcolor{link}{\ref*{#1}}}%
}
\newcommand{\tabref}[1]{%
  \tablename~\hyperref[#1]{\textcolor{link}{\ref*{#1}}}%
}
\newcommand{\eqrefc}[1]{%
  \equationname~\hyperref[#1]{\textcolor{link}{\ref*{#1}}}%
}
\newcommand{\secref}[1]{%
  \sectionname~\hyperref[#1]{\textcolor{link}{\ref*{#1}}}%
}
\newcommand{\appendixref}[1]{%
  \appendixname~\hyperref[#1]{\textcolor{link}{\ref*{#1}}}%
}

\newcommand{\tablestyle}[2]{\setlength{\tabcolsep}{#1}\renewcommand{\arraystretch}{#2}\centering\footnotesize}
\renewcommand{\paragraph}[1]{\textbf{#1.}}
\newcolumntype{x}[1]{>{\centering\arraybackslash}p{#1pt}}
\newcolumntype{y}[1]{>{\raggedright\arraybackslash}p{#1pt}}
\newcolumntype{z}[1]{>{\raggedleft\arraybackslash}p{#1pt}}
\setlist{nosep}
\newcommand{\eg}{\emph{e.g}.}
\newcommand{\ie}{\emph{i.e}.}

\usepackage{subcaption}
\usepackage{float}

\newcommand{\vs}{\emph{vs}.\ }
\usepackage{multirow}
\newlength\savewidth\newcommand\shline{\noalign{\global\savewidth\arrayrulewidth
  \global\arrayrulewidth 1pt}\hline\noalign{\global\arrayrulewidth\savewidth}}
\title{\centering \fontsize{18}{24}\selectfont{
\benchmarkname: A Challenging and Visually Diverse Multimodal Reasoning Benchmark}
}

\newcommand{\captiontitle}[1]{\textbf{#1.}}
\definecolor{wbred}{HTML}{FF8988}
\definecolor{wborange}{HTML}{FECC81}
\definecolor{wbyellow}{HTML}{F5DC51}
\definecolor{wbblue}{HTML}{6098FF}
\definecolor{wbgreen}{HTML}{77B25D}
\definecolor{wbpurple}{HTML}{B28CFF}
\definecolor{wbgray}{HTML}{9AA0A6}
\definecolor{red}{HTML}{FF8988}
\definecolor{orange}{HTML}{FECC81}
\definecolor{yellow}{HTML}{F5DC51}
\definecolor{blue}{HTML}{6098FF}
\definecolor{green}{HTML}{77B25D}
\definecolor{purple}{HTML}{B28CFF}
\definecolor{gray}{HTML}{9AA0A6}
\newcommand{\livingthings}{\textcolor{red}{\textbf{Living Things}}}
\newcommand{\objects}{\textcolor{orange}{\textbf{Objects}}}
\newcommand{\scenes}{\textcolor{yellow}{\textbf{Scenes}}}
\newcommand{\digital}{\textcolor{blue}{\textbf{Digital World}}}
\newcommand{\academics}{\textcolor{green}{\textbf{Academics}}}
\newcommand{\ocr}{\textcolor{purple}{\textbf{Documents, Charts, \& Tables}}}
\newcommand{\agents}{\textcolor{gray}{\textbf{Agents}}}
\newcommand{\alldomains}{\livingthings, \objects, \scenes, \digital, \academics, \ocr, and \agents}
\newcommand{\benchmarkname}{\mbox{WorldBench}}
\newcommand{\mmmu}{MMMU}
\newcommand{\mega}{MEGA-Bench}
\newcommand{\mme}{MME}
\newcommand{\mmbench}{MMBench}
\newcommand{\seedbench}{SEED-Bench-2}
\newcommand{\vqa}{VQAv2}
\newcommand{\mmstar}{MMStar}
\newcommand{\mmt}{MMT-Bench}

\author{
    \vspace{.2cm}
    \parbox{\textwidth}{\centering
        \Authfont
        Yida Yin\textsuperscript{1*} \hspace{.1em}
        Harish Krishnakumar\textsuperscript{1*} \hspace{.1em}
        Chung Peng Lee\textsuperscript{1} \hspace{.1em}
        Boya Zeng\textsuperscript{1} \hspace{.1em}
        Wenhao Chai\textsuperscript{1} \\
        \Authfont
        \vspace{-.5em}
        Shengbang Tong\textsuperscript{2} \hspace{.1em}
        Wenhu Chen\textsuperscript{3} \hspace{.1em}
        Hu Xu\textsuperscript{4} \hspace{.1em}
        Xingyu Fu\textsuperscript{1} \hspace{.1em}
        Gabriel Sarch\textsuperscript{1} \\
        \Authfont
        \vspace{.5em}
        Aleksandra Korolova\textsuperscript{1} \hspace{.1em}
        Zhuang Liu\textsuperscript{1\dag}
    }
    \\
    \vspace{.3cm}
    {\normalfont\fontsize{11}{15}\selectfont
    {\textsuperscript{1}Princeton University}\hspace{.1cm}}
    {\normalfont\fontsize{11}{15}\selectfont {\textsuperscript{2}NYU}\hspace{.1cm}}
    {\normalfont\fontsize{11}{15}\selectfont
    {\textsuperscript{3}University of Waterloo}}
    {\normalfont\fontsize{11}{15}\selectfont
    {\textsuperscript{4}Meta, FAIR}}\\
    \vspace{.3cm}
    $\vcenter{\hbox{\includegraphics[height=1.2em]{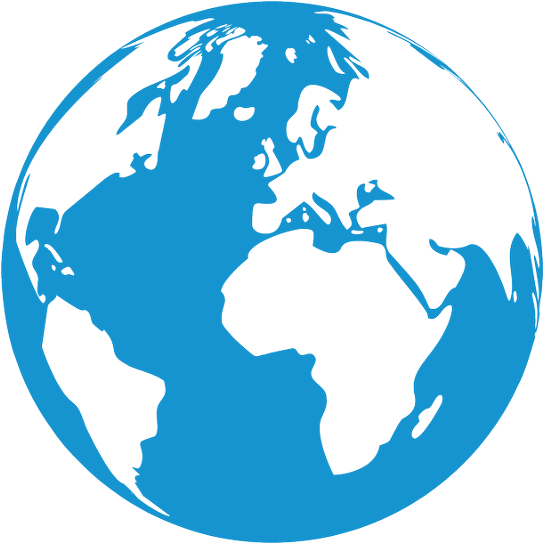}}}$\hspace{0.3em}
    \texttt{Website:} \url{https://worldbench-vl.github.io/}
    \vspace{0.2cm} \\ 
    \raisebox{-0.4ex}{\includegraphics[height=1em]{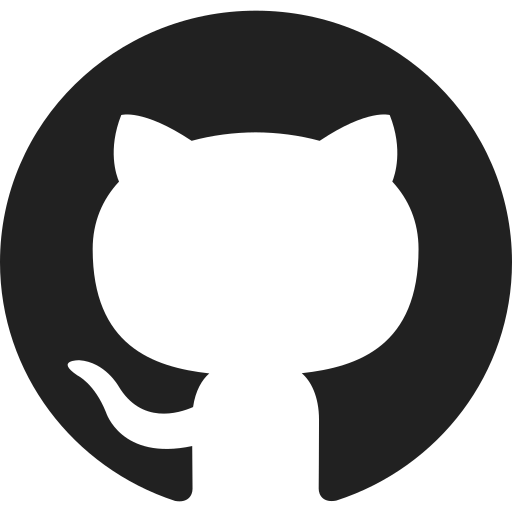}}\hspace{0.3em}\href{https://github.com/zlab-princeton/WorldBench}{\texttt{Code}} 
    \hspace{0.2cm}
    \raisebox{-0.4ex}{\includegraphics[height=1em]{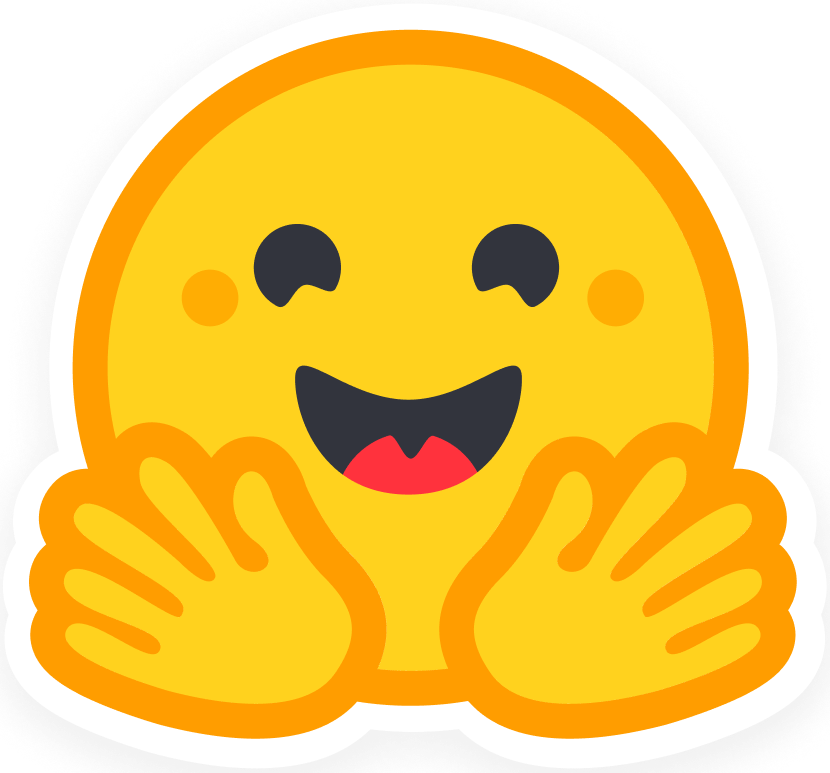}}\hspace{0.3em}\href{https://huggingface.co/datasets/zlab-princeton/WorldBench}{\texttt{Dataset}}
    \vspace{-3mm}
}

\makeatletter
\renewcommand{\abscontent}{}%
\makeatother

\newenvironment{abstractblock}{%
  {\centering\large\bfseries\sffamily Abstract\par}
  \vspace{0.2em}
  \begin{list}{}{%
      \setlength{\leftmargin}{2em}
      \setlength{\rightmargin}{2em}
      \setlength{\topsep}{0pt}
      \setlength{\parsep}{0pt}
  }
  \item[]
}{%
  \end{list}
  \par\normalfont\vspace{1em}
}

\makeatletter
\fancypagestyle{firststyle}{
  \fancyhead[L]{}
  \fancyhead[C]{}
  \fancyhead[R]{}
  \fancyfoot[L]{\footerfont \the\correspondingauthor}
  \fancyfoot[R]{}
}
\makeatother
\makeatletter
\DeclareRobustCommand\bfseries{%
  \not@math@alphabet\bfseries\mathbf
  \fontseries\bfdefault\selectfont
  \sffamily
}
\makeatother
\DeclareTextFontCommand{\textbf}{\bfseries\sffamily}

\begin{document}

\begingroup
\makeatletter
\let\raggedright\centering
\makeatother

\maketitle
\endgroup

\begingroup
\renewcommand\thefootnote{}
\footnotetext{\fontsize{9}{12}\selectfont
\hspace*{-1.8em}
\begin{tabular}{@{}r@{\ }l@{}}
$^*$ & Equal Contribution\\[-0.2em]
$^\dag$ & Corresponding Author
\end{tabular}
}
\addtocounter{footnote}{-1}
\endgroup
\newcommand{\abstractcontent}{%
In real-world applications, models are expected to perform reliably across diverse settings. Yet, many existing multimodal benchmarks expand task types without capturing the visual diversity needed to handle open-ended visual inputs. We present {\benchmarkname}, a challenging and visually diverse reasoning benchmark to evaluate Multimodal Large Language Models (MLLMs). We build a taxonomy of thousands of visual concepts across multiple domains (\eg, living things). Guided by this taxonomy, we curate a broad collection of images from search engines and existing datasets to comprehensively represent the visual world. Through structured trial-and-error, we manually design challenging questions that frontier MLLMs fail to answer. On quantitative and human evaluations, {\benchmarkname} achieves higher visual diversity than any existing diverse benchmark. Evaluating 15 MLLMs on {\benchmarkname} reveals weaknesses in visual understanding: even the strongest model reaches only 64.0\% accuracy, while some models perform marginally above chance-level. We hope our work highlights the importance of visual diversity in building multimodal benchmarks.
}
\newcommand{\teaserfigure}[2]{%
    \begin{figure}[#1]
        \centering
        \includegraphics[width=#2]{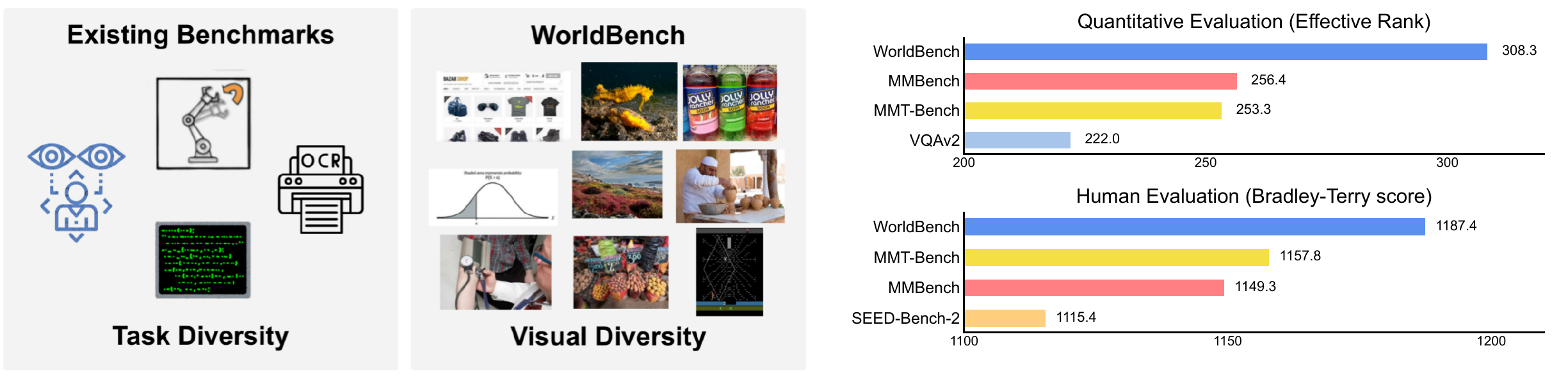}
        \caption{\captiontitle{From task diversity to visual diversity} While existing multimodal benchmarks emphasize task diversity, {\benchmarkname} focuses on visual diversity. By covering images from different visual domains, {\benchmarkname} achieves the strongest overall visual-diversity results in both quantitative and human evaluations, while remaining highly challenging for current MLLMs.}
    \end{figure}%
}
\newcommand{\linkstablestyleone}{%
    \begin{center}
        \small
        \renewcommand{\arraystretch}{1.2}
        \begin{tabular}{rll}
            \worldwideweb & \textbf{Website} & \url{https://www.test.com/}\\
            \github & \textbf{Code} & \url{https://github.com/test}\\
            \hf & \textbf{Data} & \url{https://huggingface.com}
        \end{tabular}
    \end{center}%
}

\newcommand{\linkstablestyletwo}{%
    \noindent\small
    \textbf{Website:} \url{https://www.test.com/}\\
    \textbf{Code:}    \url{https://github.com/test}\\
    \textbf{Data:}    \url{https://huggingface.com}
}

\makeatletter
\ifthenelse{\equal{\templateoption}{option1}}{
    \vspace{-0.1cm}
    
    \teaserfigure{!ht}{\textwidth}

    \vspace{0.3cm}

    \begin{abstractblock}
    \abstractcontent
    \end{abstractblock}

    \newpage
}{
    \ifthenelse{\equal{\templateoption}{option2}}{
        \vspace{-0.3cm}
        
        \begin{abstractblock}
        \abstractcontent
        \end{abstractblock}

        \vspace{0.0cm}

        \teaserfigure{!bh}{\textwidth}

        \newpage
    }{
        \begin{abstractblock}
        \abstractcontent
        \end{abstractblock}

        \vspace{0.5cm}

        \vspace{-0.5cm}
    }
}
\makeatother

\section{Introduction}

Comprehensively evaluating multimodal Large Language Models (MLLMs) requires benchmarks that capture the richness and diversity of the visual world. In recent years, many benchmarks~\citep{yue2024mmmu,fu2023mme,liu2024mmbench,li2023seed} have been introduced as \emph{standalone test suites} to evaluate models across different capabilities. Most of them follow a \emph{task-centric} construction paradigm: they first design a set of task categories (\eg, object recognition, OCR) and then build questions to cover each category.

\begin{figure}[t]
    \centering
    \includegraphics[width=\linewidth]{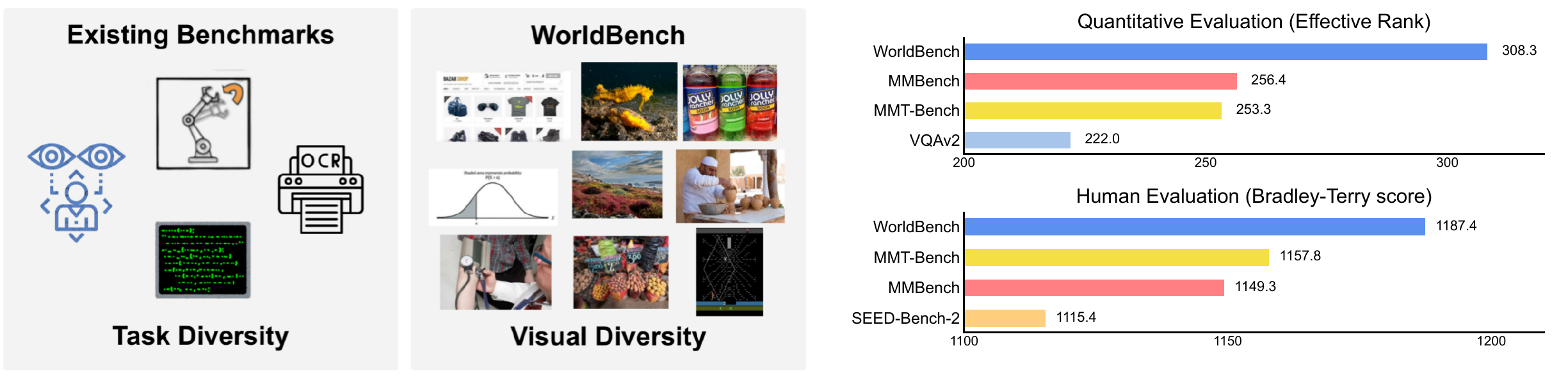}
    \caption{\captiontitle{From task diversity to visual diversity} While existing multimodal benchmarks emphasize task diversity, {\benchmarkname} focuses on visual diversity. By covering images from different visual domains, {\benchmarkname} achieves the strongest overall visual-diversity results in both quantitative and human evaluations, while remaining highly challenging for current MLLMs.}
    \label{fig:teaser}
\end{figure}

However, this task-oriented approach overlooks the visual diversity of the images. The variety of the images determines what types of questions can be asked and, therefore, which model abilities can be evaluated. We introduce {\benchmarkname}, a reasoning benchmark that assesses multimodal understanding using a diverse image set paired with questions that are intuitive for humans yet challenging for models. As shown in~\figref{fig:teaser}, {\benchmarkname} stands apart from existing diverse benchmarks by challenging models to \emph{see the whole world}.

Our argument has three parts. First, we construct {\benchmarkname} around visual diversity rather than task diversity, using a taxonomy to sample broadly from the visual world. Second, we validate that this construction produces a more visually diverse image set, using both embedding-based diversity metrics and human judgments. Third, we show that the resulting benchmark is challenging for current MLLMs, revealing visual-understanding failures that are not well captured by existing task-centric benchmarks.

To comprehensively capture the visual world, we first construct a large-scale taxonomy containing thousands of fine-grained visual concepts across 7 visual domains. This process is semi-automated with a Large Language Model (LLM) and involves light human effort. This taxonomy guides the curation of high-quality, visually diverse images from web search engines and existing datasets. Following~\citet{lin2014microsoft}, we prioritize non-iconic images that depict rich contextual scenes rather than object-centric views. For each image, we design a question through a structured trial-and-error process, iteratively refining it until at least one frontier MLLM fails to answer. Overall, {\benchmarkname} contains 2,000 carefully curated questions.

To demonstrate that {\benchmarkname} is visually more diverse than existing benchmarks, we perform both quantitative and human evaluations. Specifically, we measure visual diversity using the effective rank and participation ratio of the feature covariance matrix computed from image embeddings extracted by pre-trained vision encoders. Across different encoders, {\benchmarkname} consistently ranks among the top two benchmarks. We further conduct a user study in which participants compare two image panels sampled from different benchmarks and vote for the one they perceive as more diverse. We aggregate these pairwise votes into a global ranking using the Bradley--Terry model~\citep{bradley1952rank}, where {\benchmarkname} obtains the highest overall human-rated diversity score.

We evaluate 15 MLLMs on {\benchmarkname} and find it very challenging for current models. Among all models, Gemini-3.1-Pro~\citep{gemini31} achieves the highest accuracy of 64.0\% averaged across all domains, while the best-performing open-source model, Qwen3.5-VL-27B~\citep{qwen35}, attains 56.6\%. Several models remain only moderately above chance-level performance.

Moreover, we find that no model attains over 75\% accuracy on questions in any domain. Current models fail on many simple cases. They often struggle with fine-grained perception (\eg, counting) and make ungrounded inferences without closely looking at the image. We also find that while reasoning models outperform non-reasoning variants, increasing the number of reasoning tokens does not always improve performance on {\benchmarkname}, and this trend is observed across all domains.

\begin{figure*}[t]
    \centering
    \includegraphics[width=.90\linewidth]{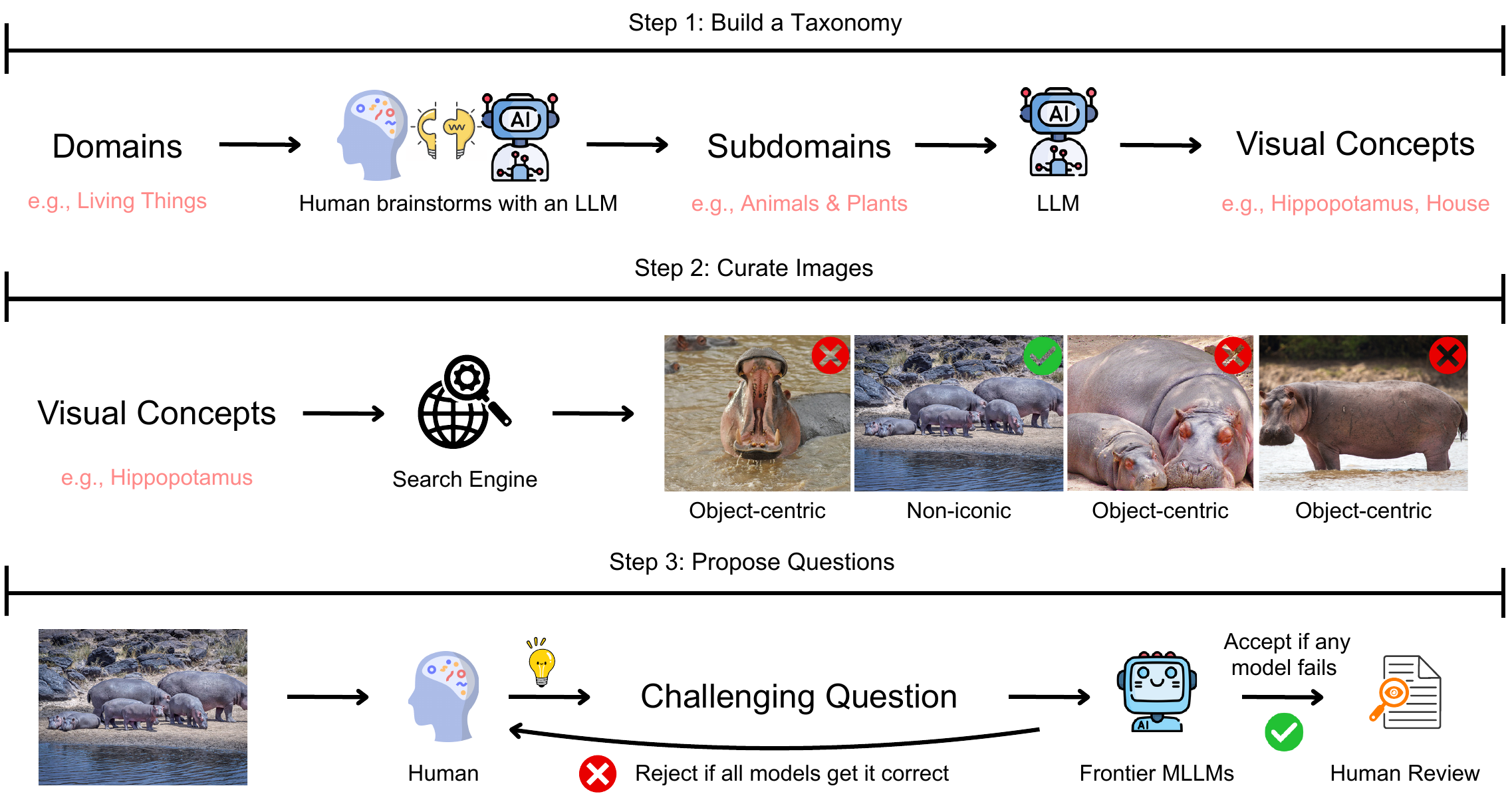}
    \caption{\captiontitle{Three-step procedure to build {\benchmarkname}} \textbf{Top:} We construct a comprehensive taxonomy by defining a few high-level domains, expanding each into multiple subdomains, and then further into fine-grained concepts. 
    \textbf{Middle:} We curate a high-quality image for each concept from search engines, prioritizing non-iconic and less object-centric images.
    \textbf{Bottom:} For each image, we design a challenging question that at least one frontier model answers incorrectly.
    }
    \label{fig:benchmark_construction}
\end{figure*}
\begin{figure*}[t]
    \centering

    \centering
    \begin{overpic}[width=\linewidth]{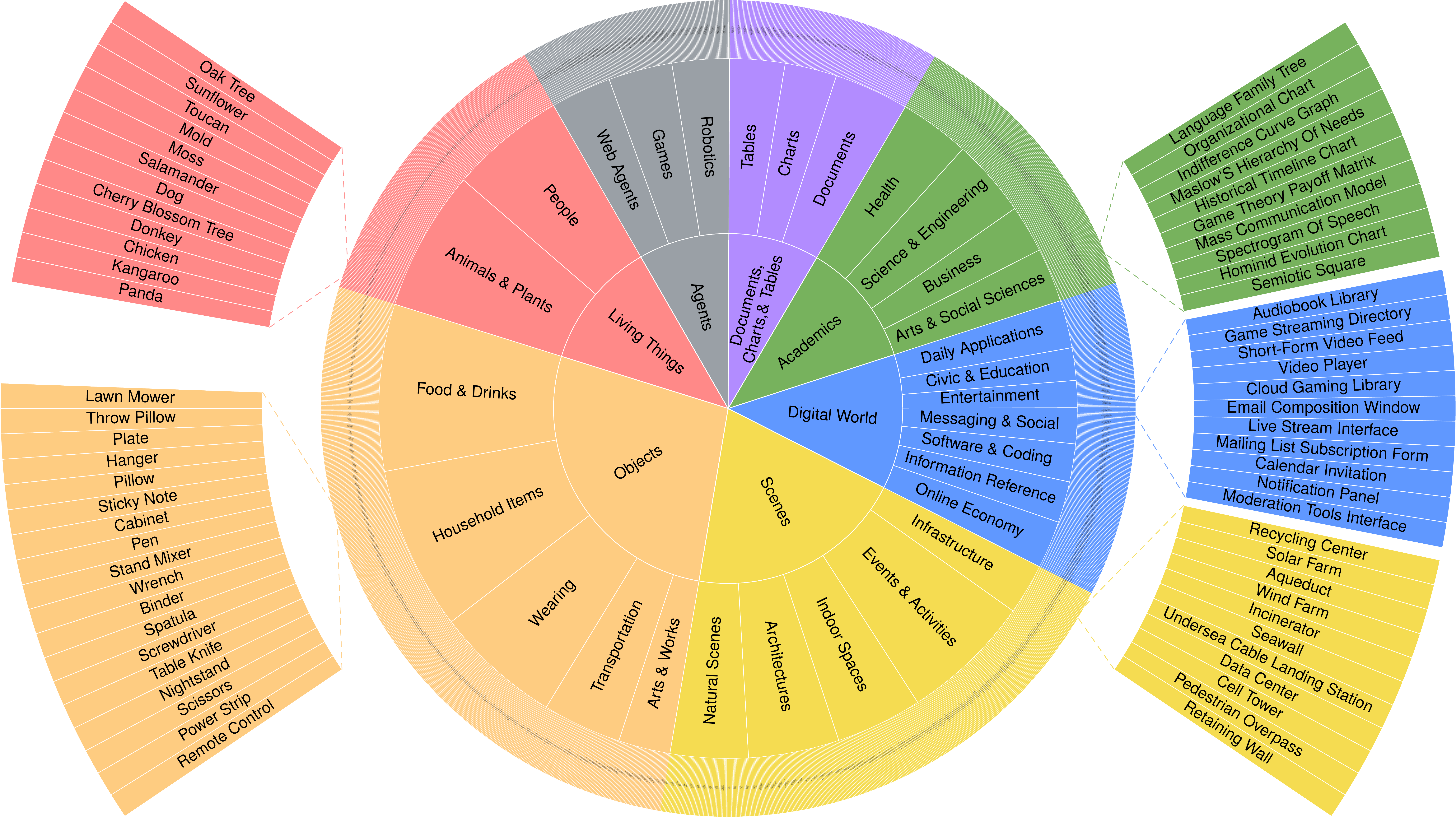}
        \put(70.8,32.9){\includegraphics[width=0.09\linewidth]{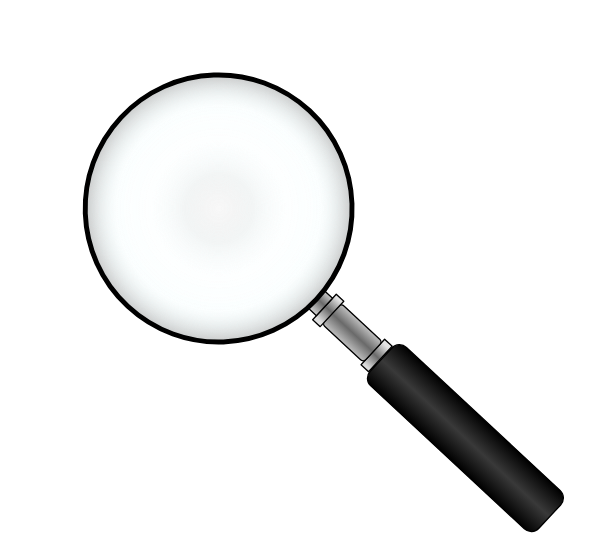}}
        \put(69.3,9.9){\includegraphics[width=0.09\linewidth]{figures/magnifier.png}}
        \put(72.8,22.9){\includegraphics[width=0.09\linewidth]{figures/magnifier.png}}
        \put(20.5,30.9){\includegraphics[width=0.08\linewidth]{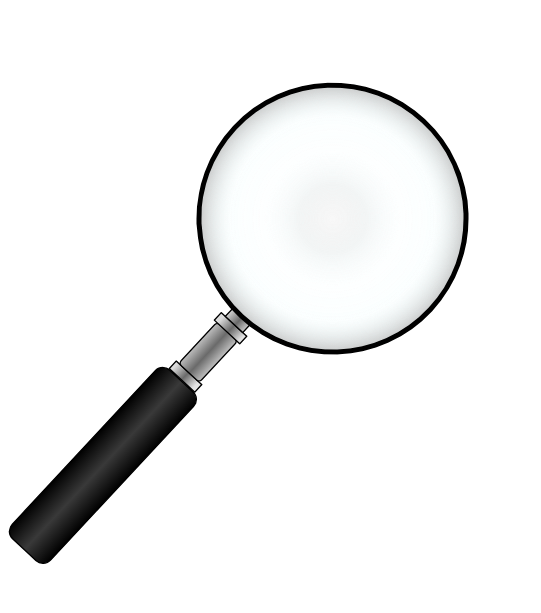}}
        \put(20.1,15.5){\includegraphics[width=0.08\linewidth]{figures/magnifier_reverse.png}}
    \end{overpic}

    \caption{\captiontitle{Taxonomy of {\benchmarkname}} 
    We construct a large-scale taxonomy covering 2,000 visual concepts across 7 domains: \livingthings, \objects, \scenes, \digital, \academics, \ocr, and \agents. Zoom in to view fine-grained concepts on the outermost ring.}
    \label{fig:taxonomy}
\end{figure*}
\section{\benchmarkname}
Existing multimodal benchmarks~\citep{fu2023mme, mmtbench, chen2025megabench} often emphasize task diversity, but rarely consider the visual diversity of their images during construction. We present \benchmarkname, a new reasoning benchmark to challenge multimodal Large Language Models (MLLMs) to \emph{see the whole world} by evaluating them on a broad set of visually diverse images across 7 domains, including \alldomains. {\benchmarkname} comprises 2,000 questions that are challenging for frontier MLLMs but natural and intuitive for humans.

\figref{fig:benchmark_construction} illustrates our three-step procedure to build {\benchmarkname}: (1) to capture the diversity of the visual world, we construct a large-scale taxonomy covering a spectrum of visual concepts with the help of a Large Language Model (LLM); (2) we curate a diverse set of images by selecting a high-quality sample for each concept; and (3) we manually propose difficult questions that frontier MLLMs answer incorrectly.

\begin{figure*}[t]

  \centering
  \begin{subfigure}[t]{0.495\linewidth}
    \centering
    \includegraphics[width=\linewidth]{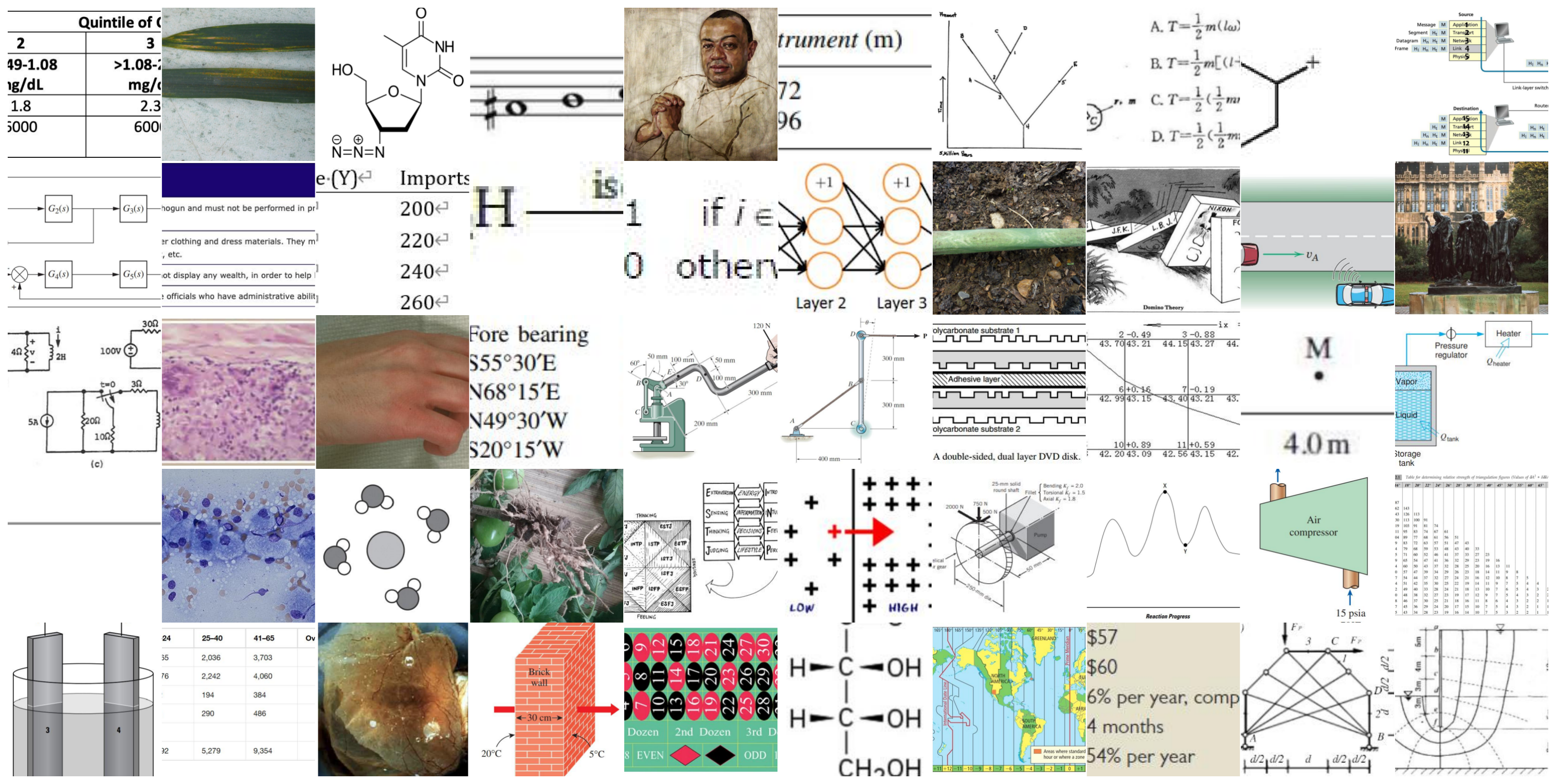}%
\caption*{\centering\mmmu\citep{yue2024mmmu}} \label{fig:mmmu}
  \end{subfigure}\hfill
  \begin{subfigure}[t]{0.495\linewidth}
    \centering
    \includegraphics[width=\linewidth]{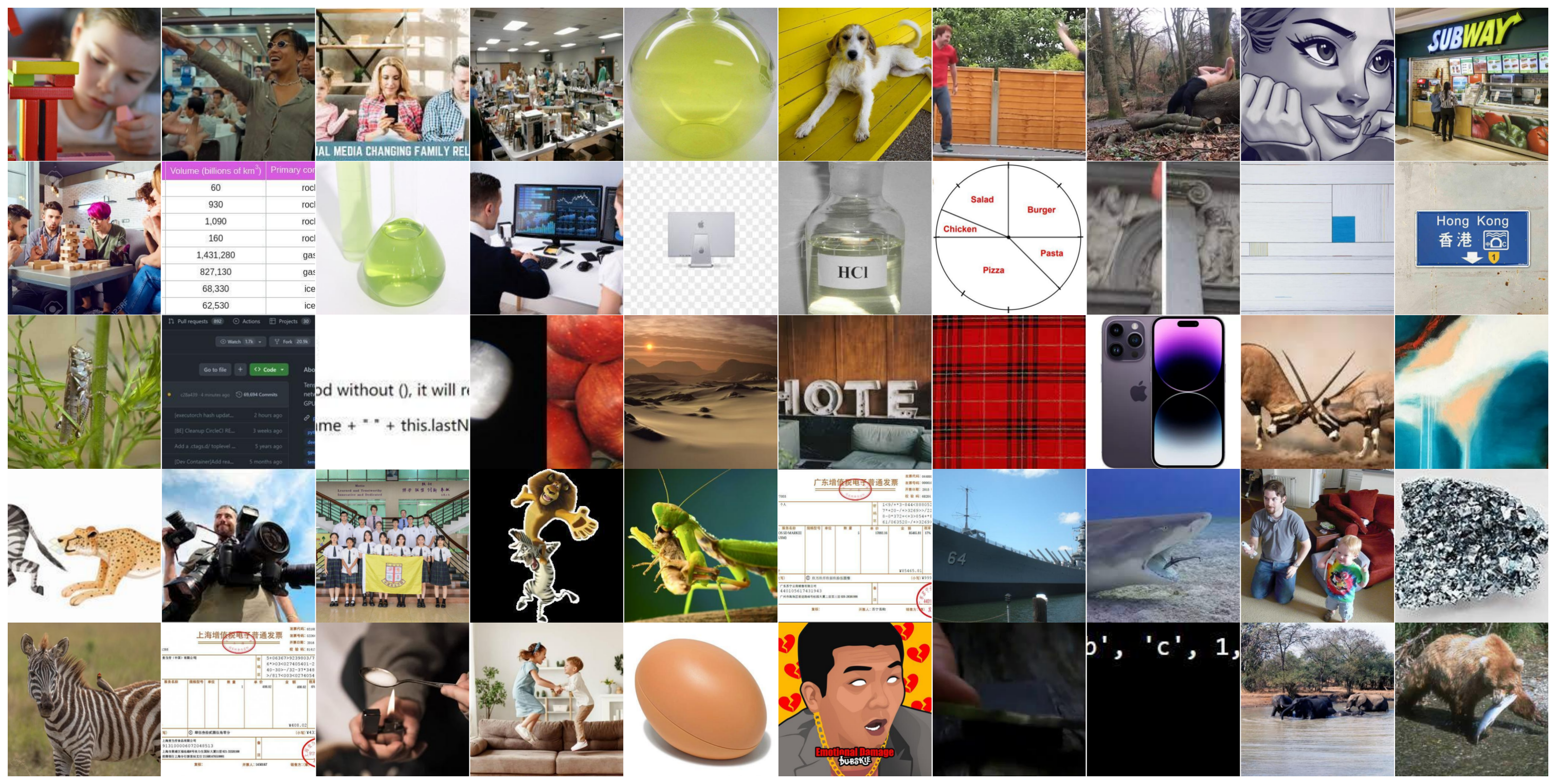}%
    \caption*{\centering\mmbench~\citep{liu2024mmbench}} \label{fig:mmbench}
  \end{subfigure}

  \begin{subfigure}[t]{0.495\linewidth}
    \centering
    \includegraphics[width=\linewidth]{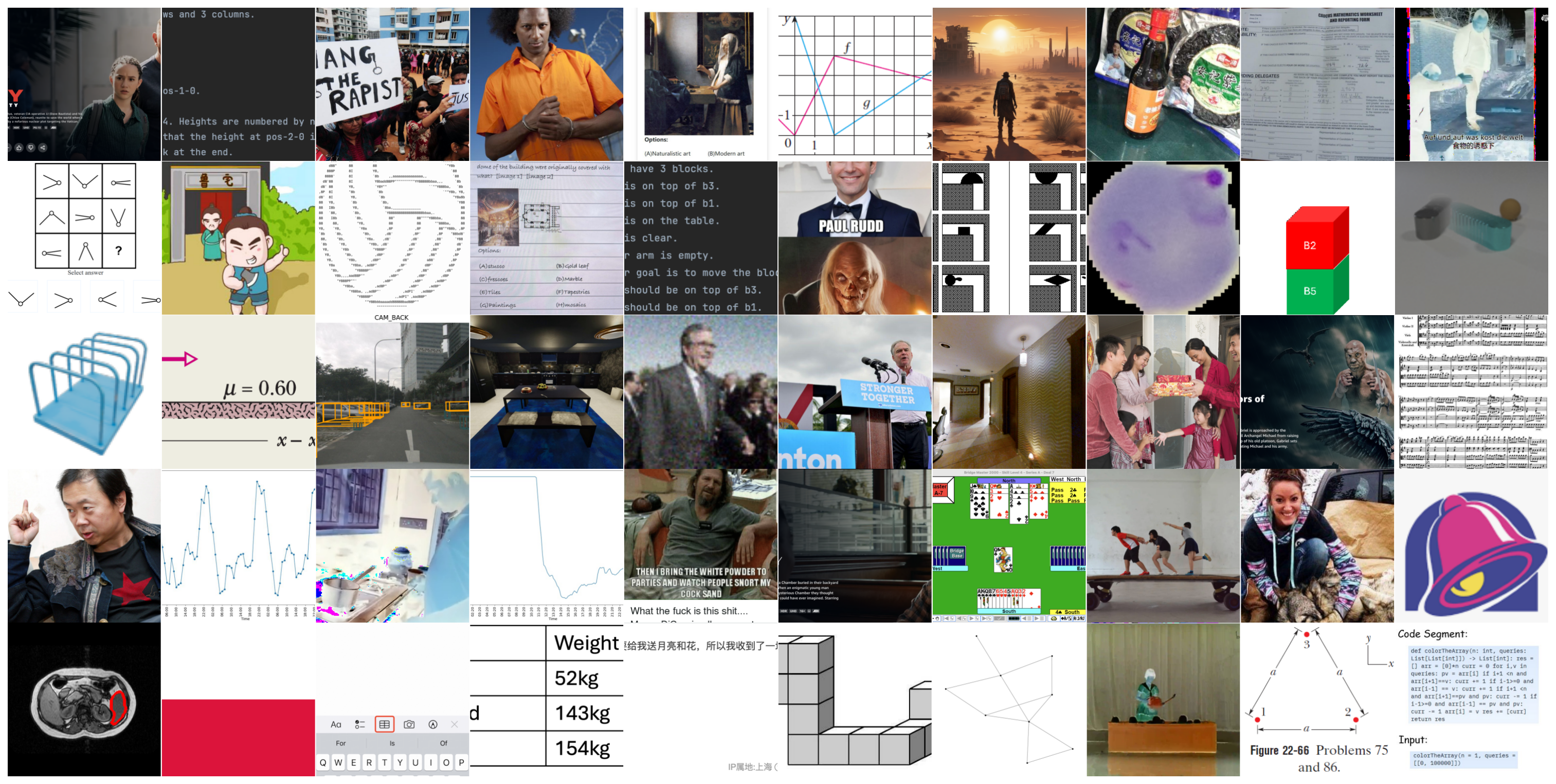}%
    \caption*{\centering\mega~\citep{chen2025megabench}} \label{fig:megabench}
  \end{subfigure}\hfill
  \begin{subfigure}[t]{0.495\linewidth}
    \centering
    \includegraphics[width=\linewidth]{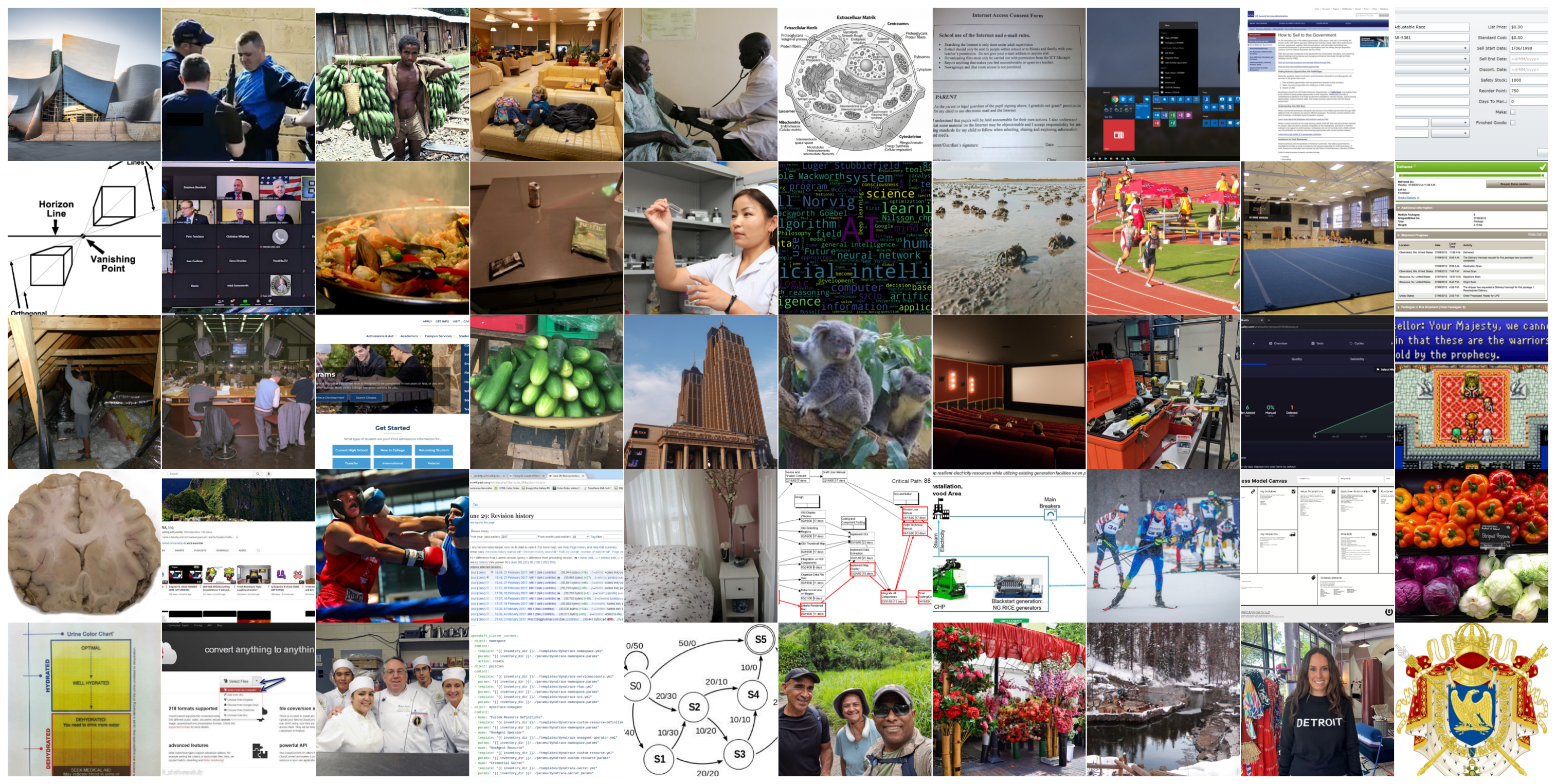}%
    \caption*{\centering\benchmarkname} \label{fig:ours}
  \end{subfigure}
  \caption{\captiontitle{Uncurated random image samples from {\benchmarkname} \vs existing multimodal benchmarks} Images in \mmbench~\citep{liu2024mmbench} are often close-up and object-centric, while those in {\mmmu} and {\mega} largely consist of charts and diagrams. In contrast, {\benchmarkname} covers a variety of visual concepts, ranging from real-world objects and scenes to web content and games. For visualization, each image in the panel is cropped to a square size.}
    
  \label{fig:image_samples}
\end{figure*}
\subsection{Building a Comprehensive Taxonomy}
\label{sec:taxonomy}

Here we discuss why existing taxonomies are insufficient to help curate diverse images and introduce ours.

\paragraph{Limitations in existing taxonomies} Taxonomies have long been used to curate image datasets~\citep{griffin2007caltech, Deng2009} in computer vision. Many rely on large-scale, off-the-shelf semantic backbone (\ie, WordNet~\citep{miller1995wordnet}) or manually construct hierarchies with a few hundreds common object categories~\cite{everingham2010pascal, lin2014microsoft}. These taxonomies are predominated by tangible, real-world entities (\eg, animals, vehicles) but overlook many forms of web-native visual content that are increasingly important for modern MLLM applications. For instance, ImageNet allocates $>$120 of its 1,000 categories to fine-grained dog breeds but covers no digital imagery (\eg, webpages, screenshots, games).

In addition, there are other publicly available ones used by industries or organizations. They are designed to meet specific practical needs (\eg, structuring commercial product listings~\citep{google_trend_categories, google_product_taxonomy}, or organizing knowledge~\citep{wikipedia_categories, wikidata}). However, we find that the concepts in these taxonomies are often limited to narrow domains and do not generalize to the broad visual diversity that we are interested in.

\paragraph{LLM-guided taxonomy construction} Manually constructing a diverse taxonomy is very time-consuming, so we use an LLM to assist in developing taxonomy with \emph{light human-in-the-loop effort}. We first define a few high-level visual domains---\livingthings, \objects, \scenes, \digital, \academics, \ocr, and \agents---to serve as the backbone of our taxonomy.

For all domains except {\agents}, we prompt an LLM to propose candidate subdomains. We manually review them to prune redundant ones and merge semantically similar ones. This generate-refinement cycle is repeated multiple times until no major new subdomains appear. We then prompt the LLM to expand each subdomain into fine-grained visual concepts.

For {\agents}, we manually define three subdomains---\textcolor{gray}{\textbf{Robotics}}, \textcolor{gray}{\textbf{Games}}, and \textcolor{gray}{\textbf{Web Agents}}. For \textcolor{gray}{\textbf{Web Agents}} and \textcolor{gray}{\textbf{Games}}, we follow the same LLM-assisted taxonomy construction procedure above. But for \textcolor{gray}{\textbf{Robotics}}, we found there are fewer first-person images capturing robotic environments online. We therefore source such images from the robotics domain in AgentVQA~\citep{anonymous2025agentvqa}, a unified benchmark for agentic visual understanding. We then apply an MLLM to create a taxonomy of about 60 leaf concepts that cover the majority of robotic scenarios in this dataset.

\figref{fig:taxonomy} visualizes our taxonomy with 2,000 visual concepts in 7 visual domains. The concepts range from \textcolor{red}{\textbf{crab}} and \textcolor{red}{\textbf{snake}} in \textcolor{red}{\textbf{Animals \& Plants}} to \textcolor{yellow}{\textbf{comedy show}} and \textcolor{yellow}{\textbf{hiking}} in \textcolor{yellow}{\textbf{Events \& Activities}}. They also cover \textcolor{blue}{\textbf{shopping cart interface}} and \textcolor{blue}{\textbf{product review}} in \textcolor{blue}{\textbf{Online Economy}}, \textcolor{green}{\textbf{isometric drawing}} and \textcolor{green}{\textbf{supply and demand curve}} in \textcolor{green}{\textbf{Science \& Engineering}}, as well as \textcolor{gray}{\textbf{booking a flight}} and \textcolor{gray}{\textbf{installing an app}} in \textcolor{gray}{\textbf{Web Agents}}.

\subsection{Curating Images from Taxonomy}
\label{sec:curate_image}

Since the taxonomy defines the target visual concepts for our benchmark, we then collect high-quality images corresponding to these concepts from reliable sources. Below we discuss our image curation method and criteria, followed by a qualitative comparison between our images and those in existing diverse benchmarks.

\paragraph{Collecting diverse images from search engines} Following classic vision datasets~\citep{Deng2009,lin2014microsoft}, we use search engines as the primary source of images. For each visual concept, we query search engines (\eg, Google and Bing) to obtain a large pool of candidate images, typically a few hundred images per query. We then select the image that best matches the target concept. If no such image exists in the initial pool, we query more images from search engines until a good one is found.

We observe that search engines tend to return object-centric, close-up images, particularly for real-world domains (\eg, \livingthings\ and \objects). We avoid overusing such images in our benchmark, as they would greatly reduce the visual diversity. Similar to~\citet{lin2014microsoft}, we prioritize non-iconic images (or non-canonical perspectives~\citep{palmer1981cannonical}) with richer contexts and possibly containing multiple objects.

The above procedure applies to all subdomains except \textcolor{gray}{\textbf{Robotics}} in \agents. Because the taxonomy for this subdomain is derived from robot-centric images in AgentVQA~\citep{anonymous2025agentvqa}, we use an MLLM to select one representative image for each visual concept directly from this dataset, followed by human verification. This approach largely preserves visual diversity of robot-centric images in the original dataset while avoiding choosing many similar images from the same source.

\paragraph{Comparing to images in other diverse benchmarks} In total, {\benchmarkname} consists of 2,000 images. \figref{fig:image_samples} shows uncurated random samples from {\benchmarkname} along with those from existing diverse benchmarks. {\mmmu}~\citep{yue2024mmmu} and {\mega}~\citep{chen2025megabench} feature textbook-style charts and diagrams, while images in {\mmbench}~\citep{liu2024mmbench} often depict a single centered object with minimal context. In contrast, {\benchmarkname} offers greater visual diversity and a more balanced distribution across concepts. We will quantitatively show that {\benchmarkname} is visually more diverse than existing diverse benchmarks in~\secref{sec:diversity}.

\begin{figure*}[t]
    \centering
    \includegraphics[width=\linewidth]{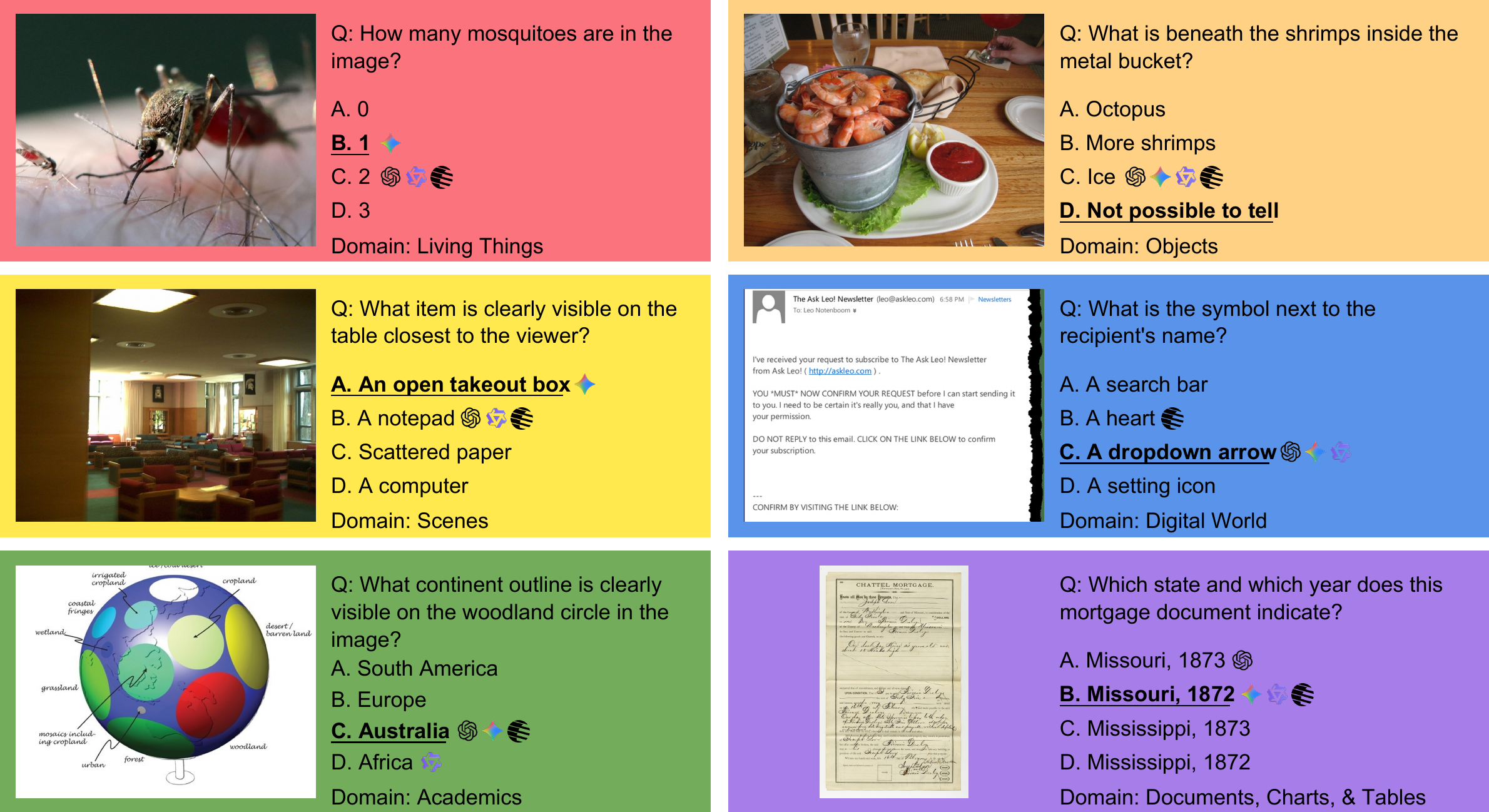}
    \caption{\captiontitle{Question samples from {\benchmarkname}} We show representative questions with answers selected by frontier MLLMs. Ground-truth answers are highlighted in \textbf{\underline{bold and underline}}.}
    \label{fig:question}
\end{figure*}
\subsection{Proposing Challenging Questions}
\label{sec:questions}

Now that we have curated a diverse set of images across domains, we describe how we design difficult questions for frontier MLLMs and perform human review to control annotation error. We also highlight the key features that set {\benchmarkname} apart from other challenging multimodal benchmarks.

\paragraph{Designing challenging questions against models} We adopt a structured trial-and-error procedure to design difficult questions that require models to reason carefully before providing answers. We draft an initial question for each image and evaluate it on several frontier models (\ie, GPT-5.4-Thinking (low)~\citep{gpt54}, Gemini-3.1-Pro~\citep{gemini31}, Qwen3.5-VL-Plus-Instruct~\citep{qwen35}, and Kimi-K2.5~\citep{team2026kimi}). The question is iteratively revised until at least one model answers incorrectly. During benchmark construction, newer and more capable models are released, so different questions are evaluated against different sets of models; \tabref{tab:models_evaluate_against} reports these details.

\begin{table}[h]
    \centering
    \tablestyle{6pt}{1.0}
    \begin{tabular}{ll}
        \# of questions & models \\
        \shline
        321 & GPT-5-Thinking, Gemini-2.5-Flash, Qwen3-VL-Plus-Instruct \\
        352 & GPT-5-Thinking, Gemini-2.5-Flash, Qwen3-VL-Plus-Instruct, Seed1.6-Vision \\
        779 & GPT-5-Thinking, Gemini-3-Pro, Qwen3-VL-Plus-Instruct, Seed1.6-Vision \\
        542 & GPT-5.4-Thinking, Gemini-3.1-Pro, Qwen3.5-VL-Plus-Instruct, Kimi-K2.5
        
    \end{tabular}
    \caption{\captiontitle{Number of questions evaluated against each set of frontier models} Since newer and more capable models are released during benchmark construction, different questions are evaluated against different sets of models.}
    \label{tab:models_evaluate_against}
\end{table}

We place only one constraint in question design: each question must be a four-option multiple-choice question. Note that the question is not required to be directly tied to the visual concept used to retrieve the image. We also ask annotators to optionally provide a concise explanation of the correct answer for questions they consider challenging and require substantial reasoning to arrive at the final answer. \appendixref{appendix:annotation_interface} illustrates our question annotation interface, along with the tools provided to annotators in designing questions.

\paragraph{Controlling annotation error} Frontier models may fail to answer certain questions not because the questions are actually difficult, but due to bad annotation (\eg, ambiguous wording, multiple valid answers, or incorrect descriptions of the image). To mitigate this, we conduct several rounds of manual review of all questions and their corresponding explanations. We refine unclear language to eliminate alternative interpretations and carefully verify that each question admits only one valid correct answer.

We also use Claude Code to improve question quality with only light human effort in the loop. First, we ask volunteers outside the annotation team to audit 100 randomly sampled questions and document any issues they observe. We then distill these issues into general principles and use them to review all questions with Claude Code. For each question, Claude Code will suggest a revision and provide an explanation for the change if it finds the original question problematic. We adopt a proposed revision only if we agree with its explanation and consider it meaningfully improves the quality. In cases where we agree that a question is problematic but Claude Code does not produce a satisfactory revision, we manually redesign the question.

\paragraph{Sample questions in {\benchmarkname}} \figref{fig:question} shows 6 representative question samples in {\benchmarkname} along with the answers chosen by frontier models. While these questions are \emph{intuitive} for humans, frontier models often answer them incorrectly. Our questions are very different from those in other challenging reasoning benchmarks. For example, ZeroBench~\citep{roberts2025zerobench} constructs its questions by chaining several independent sub-questions into a single, complex one (\eg, count the chairs, count the letters, then multiply the two). In contrast, our questions are designed to be \emph{more natural} and \emph{closer to real-world settings}.

\section{Measuring Visual Diversity of Images}
\label{sec:diversity}
We evaluate the visual diversity of the image set in {\benchmarkname} and compare it against existing multimodal benchmarks using both quantitative and human evaluations. Quantitatively, we measure visual diversity through the covariance structure of image embeddings extracted by pre-trained vision encoders. We also conduct a user study to assess how humans perceive visual diversity across different benchmarks. 

For comparison, we include widely used multimodal benchmarks that emphasize diversity, including \vqa~\citep{goyal2017making}, \mme~\citep{fu2023mme}, \mmstar~\citep{chen2024we}, \mmbench~\citep{liu2024mmbench}, \mmmu~\citep{yue2024mmmu}, \seedbench~\citep{li2023seedbench2benchmarkingmultimodallarge}, \mmt~\citep{mmtbench}, and \mega~\citep{chen2025megabench}. \appendixref{appendix:implementation_benchmark} details how we process images from each benchmark.
\begin{figure}[t]
    \centering
    \includegraphics[width=\linewidth]{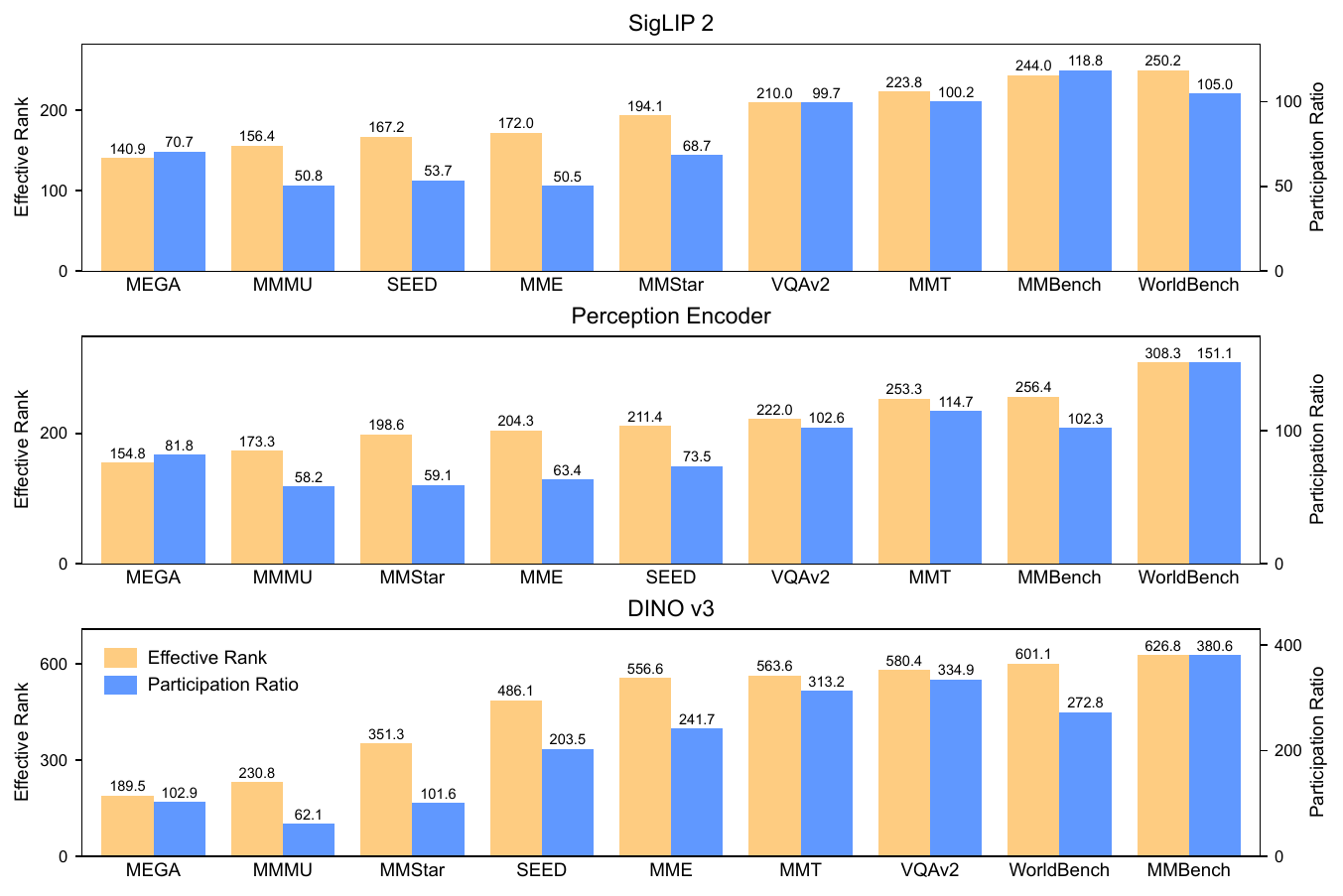}
    \caption{\captiontitle{Quantitative diversity evaluation} {\benchmarkname} often ranks first or second in effective rank~\citep{roy2007effective} and participation ratio with embeddings extracted by different pre-trained vision encoders. This indicates better diversity under varying visual representations (\ie, the distinct visual world captured by each encoder).}%
    \label{fig:embedding}
\end{figure}
\subsection{Quantitative Evaluation}
\label{sec:model_evaluation}

Vision encoders~\citep{radford2021learning, zhai2023sigmoidlosslanguageimage, oquab2023dinov2} are pre-trained on millions or even billions of images~\citep{Thomee2016, chen2023palijointlyscaledmultilinguallanguageimage, gadre2023datacomp}. As a result, they learn rich and generalizable visual representations that capture the visual world and we use them to measure visual diversity.

\paragraph{Methods}
Concretely, we randomly sample $N=1000$ images from each benchmark and extract their $\ell_2$-normalized embeddings $e_1,\dots,e_N \in \mathbb{R}^d$ using a pre-trained vision encoder. Let $\bar e$ denote their empirical mean, and define the zero-centered feature matrix:
$X = [\, e_1-\bar e,\; \dots,\; e_N-\bar e \,] \in \mathbb{R}^{d \times N}$. We then compute the sample covariance matrix $C = \frac{1}{N} XX^\top$, and denote its eigenvalues by $\lambda_1,\dots,\lambda_d \ge 0$.

To measure visual diversity, we use \emph{effective rank}\footnote{This is equivalent to the Vendi Score~\citep{friedman2022vendi} with cosine similarity.}~\citep{roy2007effective}:
\[
\mathrm{ER}(C) = \exp\!\left(
    - \sum_{i=1}^d p_i \log p_i
\right),
p_i = \frac{\lambda_i}{\sum_{j=1}^d \lambda_j}
\]
and \emph{participation ratio}:
\[
\mathrm{PR}(C) = \frac{\left(\sum_{i=1}^d \lambda_i\right)^2}{\sum_{i=1}^d \lambda_i^2},
\]
of the sample covariance matrix, following~\cite{wiedmann2025finevisionopendataneed}.

Intuitively, higher visual diversity leads to embeddings whose variance is distributed across many directions in feature space, rather than collapsing onto a few dominant ones. When this happens, the eigenvalues of $C$ have a more uniform distribution, leading to higher effective rank and participation ratio.

\paragraph{Vision encoders} For a comprehensive evaluation, we use multiple pre-trained vision encoders---SigLIP 2~\citep{tschannen2025siglip}, Perception Encoder~\citep{bolya2025perception}, and DINO v3~\citep{simeoni2025dinov3}. Each model is trained on distinct data distributions (\ie, WebLI~\citep{chen2023palijointlyscaledmultilinguallanguageimage}, MetaCLIP~\citep{xu2023demystifying}, and Instagram images $+$ ImageNet~\citep{Deng2009}), therefore offering different views of the visual world.

\paragraph{Results} \figref{fig:embedding} shows the results. {\benchmarkname} often achieves the highest or second-highest effective rank and participation ratio across three vision encoders. This indicates that, under different visual representations captured by different encoders, {\benchmarkname} maintains greater visual diversity than existing benchmarks.

We also observe that diversity rankings vary considerably across models. For example, {\mmbench} obtains the highest value in participation ratio among all benchmarks under SigLIP 2, yet drops to the fourth place under Perception Encoder. This suggests that different encoders capture distinctive visual representations. \appendixref{appendix:additional_results} discusses the limitations of using a pre-trained vision encoder as a general visual diversity metric.

\subsection{Human Evaluation}
~\label{sec:human_evaluation}

We conduct a human evaluation to compare benchmarks in visual diversity. We design a user interface that displays two image panels side by side, each containing 100 images randomly sampled from a benchmark. Users are asked to vote for the panel that \emph{is visually more diverse}. Below, we describe how we collect user responses and how we aggregate them to produce an absolute ranking of visual diversity across benchmarks. Our diversity evaluation is conceptually similar to Chatbot Arena~\citep{chiang2024chatbot}: instead of choosing a more preferred model response, users select the image panel they perceive as visually more diverse.

\paragraph{Bradley-Terry Model} We have $K$ items indexed by $i \in \{1,\dots,K\}$. The Bradley-Terry model~\citep{bradley1952rank} assigns each item a score $\theta_i \in \mathbb{R}$ that defines the probability that item $i$ is preferred over item $j$:
\begin{equation}
    \Pr(i \succ j)
    = \frac{\exp(\theta_i)}{\exp(\theta_i) + \exp(\theta_j)}
    = \sigma(\theta_i - \theta_j),
\end{equation}
where $\sigma(\cdot)$ is the sigmoid function. 

In our case, each item is a benchmark. We collect $N$ pairwise comparisons, where each comparison $n$ involves a benchmark pair $(i_n, j_n)$ and an outcome $y_n \in \{0,1\}$. The outcome $y_n = 1$ indicates that benchmark $i_n$ is considered visually more diverse than benchmark $j_n$, and $y_n = 0$ otherwise. Under the Bradley-Terry model, we estimate the score vector $\hat\theta = (\hat\theta_1,\dots,\hat\theta_K)$ by maximizing the log-likelihood over all observed comparisons:
\begin{equation}
    \mathcal{L}(\theta)
    = \sum_{n=1}^N \Big[
        y_n \log \Pr(i_n \succ j_n)
        + (1 - y_n) \log \Pr(j_n \succ i_n)
    \Big].
\end{equation}
Following~\citep{chiang2024chatbot}, the resulting scores are then linearly rescaled to Elo-style rating for readability.

\paragraph{Bootstrap confidence intervals}
To quantify uncertainty in the estimated scores, we compute bootstrap confidence intervals~\citep{diciccio1996bootstrap}. We resample the observed pairwise comparisons with replacement, refit the Bradley-Terry model on each set bootstrap sample, and record the resulting score for each benchmark. We perform 10,000 bootstrap rounds to obtain a stable empirical distribution of scores.

During bootstrapping, we uniformly resample every benchmark pair to mitigate imbalance across comparisons in our original data. For each benchmark, we report the mean bootstrap rating together with a 95\% confidence interval, defined by the 2.5th and 97.5th percentiles of its bootstrap rating distribution.

\paragraph{Users} We collected 360 pairwise comparison results from 12 volunteers participating in our study. None of them were involved in image curation or had any prior knowledge of our benchmark before the test. Each user completes 30 rounds of comparisons. We do not impose any time limits for a round or the entire test, and only recommend spending at least one minute per round.

\begin{figure}[t]
    \centering
    \includegraphics[width=.75\linewidth]{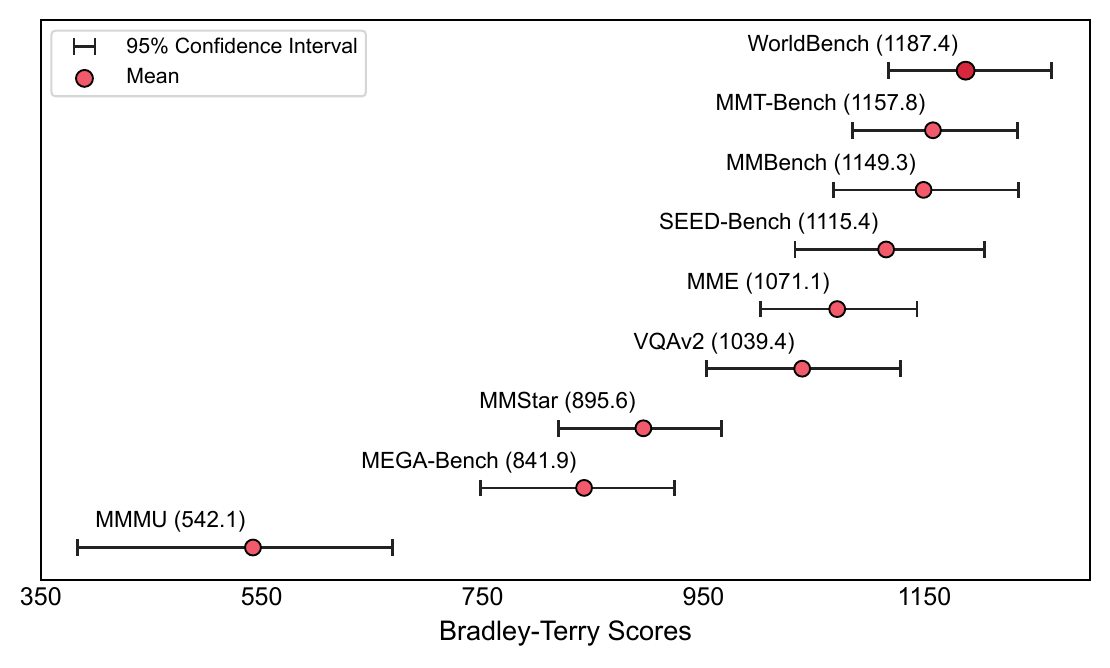}
    \caption{\captiontitle{Human evaluation on visual diversity} Bradley--Terry scores with 95\% confidence intervals estimated from pairwise human comparisons. Overall, {\benchmarkname} receives the highest human-rated diversity score.}
    \label{fig:human_evaluation}
\end{figure}

\paragraph{Results} \figref{fig:human_evaluation} visualizes Bradley--Terry scores estimated from pairwise human comparisons. {\benchmarkname} achieves the highest score, indicating that users judge its image set to be more visually diverse than those of existing benchmarks.

While our test assesses the visual diversity for a benchmark on a system level, users may rely on specific criteria to judge. Therefore, we further asked users to explain the criteria they used when selecting the more diverse image panel. Their responses commonly referenced semantic content and category breadth, the balance between charts/diagrams and real photos, visual variety, and the absence of duplicate or near-duplicate images.

Among these criteria, some are simple and objective to identify a more diverse image panel (\eg, “fewer duplicate or similar images”). Others, such as “semantic content \& category”, depend more on individual judgment.

\begin{table*}[t]
    \centering
    \tablestyle{0pt}{1.0}
    \begin{tabular}{lx{45.5}x{45.5}x{45.5}x{45.5}x{45.5}x{45.5}x{45.5}x{45.5}}
    \shline
    \textbf{Models}     &  \textbf{\textcolor{red}{Living}} & \textbf{\objects} & \textbf{\scenes} & \textbf{\textcolor{blue}{Digital}} & \textbf{\academics} & \textbf{\textcolor{purple}{DCT}} & \textbf{\textcolor{gray}{Agents}} & \textbf{Average} \\
    \Xhline{0.7pt}
    \multicolumn{9}{c}{\emph{Proprietary Models}} \\                                                                 
  GPT-5.4-Thinking (high) & 55.5 & 53.8 & 54.3 & \underline{67.3} & 60.1 & 64.5 & \textbf{63.9} & 58.2 \\
    GPT-5.4-Thinking (low)$^*$ & 51.8 & 48.4 & 50.3 & 58.1 & 53.9 & 61.6 & 58.4 & 53.0 \\
    Gemini-3.1-Pro$^*$ & \textbf{62.0} & \underline{60.7} & \textbf{61.4} & \textbf{71.4} & \textbf{64.0}
  & \textbf{73.8} & 62.0 & \textbf{64.0} \\
    Gemini-3-Flash & \underline{56.7} & \underline{60.7} & \underline{57.5} & 66.9 & \underline{63.2} &
  \underline{72.7} & \underline{63.3} & \underline{61.8} \\
    Claude-Opus-4.7 & 46.1 & 52.8 & 51.3 & 59.3 & 57.0 & 58.7 & 56.0 & 53.7 \\
    Grok-4.2 & 49.8 & 53.6 & 49.2 & 59.7 & 53.9 & 57.0 & 54.2 & 53.3 \\
    Qwen3.5-VL-Plus-Thinking & 56.3 & \textbf{62.8} & 55.0 & 58.9 & 60.1 & 60.2 & \underline{63.3} & 59.3
  \\
  Qwen3.5-VL-Plus-Instruct$^*$ & 45.7 & 49.4 & 45.1 & 55.2 & 47.8 & 49.1 & 51.2 & 48.7 \\
    \Xhline{0.7pt}
    \multicolumn{9}{c}{\emph{Open-source Models}} \\
    Qwen3.5-VL-35B-A3B & 48.6 & 52.6 & 52.2 & 50.4 & 57.0 & 59.3 & 54.2 & 52.9 \\
    Qwen3.5-VL-27B & 50.6 & 57.5 & 53.3 & 58.5 & 61.8 & 57.6 & 60.2 & 56.6 \\  
    Kimi-K2.5$^*$ & 49.0 & 52.0 & 49.7 & 56.9 & 51.3 & 59.3 & 54.8 & 52.5 \\
    GLM-4.6V & 47.8 & 41.1 & 43.4 & 38.7 & 40.4 & 42.4 & 44.6 & 42.5 \\
    Gemma-4-31B & 49.4 & 47.4 & 46.0 & 51.2 & 54.4 & 52.9 & 54.8 & 49.7 \\
    Gemma-4-E4B & 34.3 & 33.6 & 29.7 & 38.7 & 36.8 & 39.0 & 37.3 & 34.6 \\
    InternVL-3.5 & 38.4 & 42.5 & 44.6 & 39.1 & 38.6 & 42.4 & 38.6 & 41.2 \\
    \shline
    \end{tabular}
    \caption{\captiontitle{Performance of MLLMs on {\benchmarkname}} \textbf{Bold} indicates the highest value in each visual domain (column), and \underline{underline} marks the second-highest. 
    *: Models used during the question–proposal stage (\secref{sec:questions}). 
    }
    \label{tab:evaluation}
\end{table*}

\section{Model Performance}
\label{sec:model_performance}

\subsection{Settings}
\paragraph{Models} We evaluate the performance of 15 MLLMs on {\benchmarkname}. For proprietary models, we consider GPT-5.4-Thinking~\citep{gpt54}, Gemini-3.1-Pro~\citep{gemini31}, Gemini-3-Flash~\citep{gemini3}, Claude-Opus-4.7~\citep{claude47opus}, Grok-4.2~\citep{grok42}, and Qwen3.5-VL-Plus-Thinking/Instruct~\citep{qwen35}. These models are accessed through their official APIs. For open-source models, we consider Qwen3.5-VL-35B-A3B~\citep{qwen35}, Qwen3.5-VL-27B~\citep{qwen35}, Kimi-K2.5 (1T)~\citep{team2026kimi}, GLM-4.6V (106B)~\citep{glm46}, Gemma-4 (31B/E4B)~\citep{gemma4}, and InternVL3.5 (241B)~\citep{wang2025internvl3}. We run offline inference with vLLM~\citep{kwon2023efficient} for all open-source models except Kimi-K2.5, which we evaluate through its API due to its size.

\paragraph{Evaluation setup}
We use a unified prompt across all models to format the multiple choice questions, encouraging each model to provide a brief explanation before outputting its final answer choice. Final answers are extracted using regular expressions based on a fixed set of parsing rules. For generation, we use the default sampling parameters (e.g., temperature, top-k, and top-p) specified in each model’s official configuration file or API. Additional evaluation details are provided in~\appendixref{appendix:implementation_model}.

\subsection{Results}

\tabref{tab:evaluation} reports the performances of all evaluated models on {\benchmarkname}. The results show that our benchmark presents a great challenge for both proprietary and open-source models. The strongest proprietary model (\ie, Gemini-3.1-Pro) achieves an average accuracy of 64.0\% across all domains, while the best open-source model (\ie, Qwen3.5-VL-27B) attains 56.6\%. Interestingly, a few open-source models (\eg, Qwen3.5-VL-27B) outperform proprietary models (\eg, Claude-Opus-4.7 and Grok-4.2) that are often considered more capable.

No single model achieves accuracy $>$75\% on any domain. The performance gap between models within each domain is also substantial. For instance, Gemini-3.1-Pro reaches 73.8\% accuracy on {\ocr}, while Gemma-4-E4B reaches only 39.0\% on the same domain, only moderately above chance-level performance.

\subsection{Analysis}
\label{sec:analysis}

\begin{figure}[t]
    \centering
    \begin{minipage}[t]{0.49\linewidth}
        \centering
        \includegraphics[width=\linewidth]{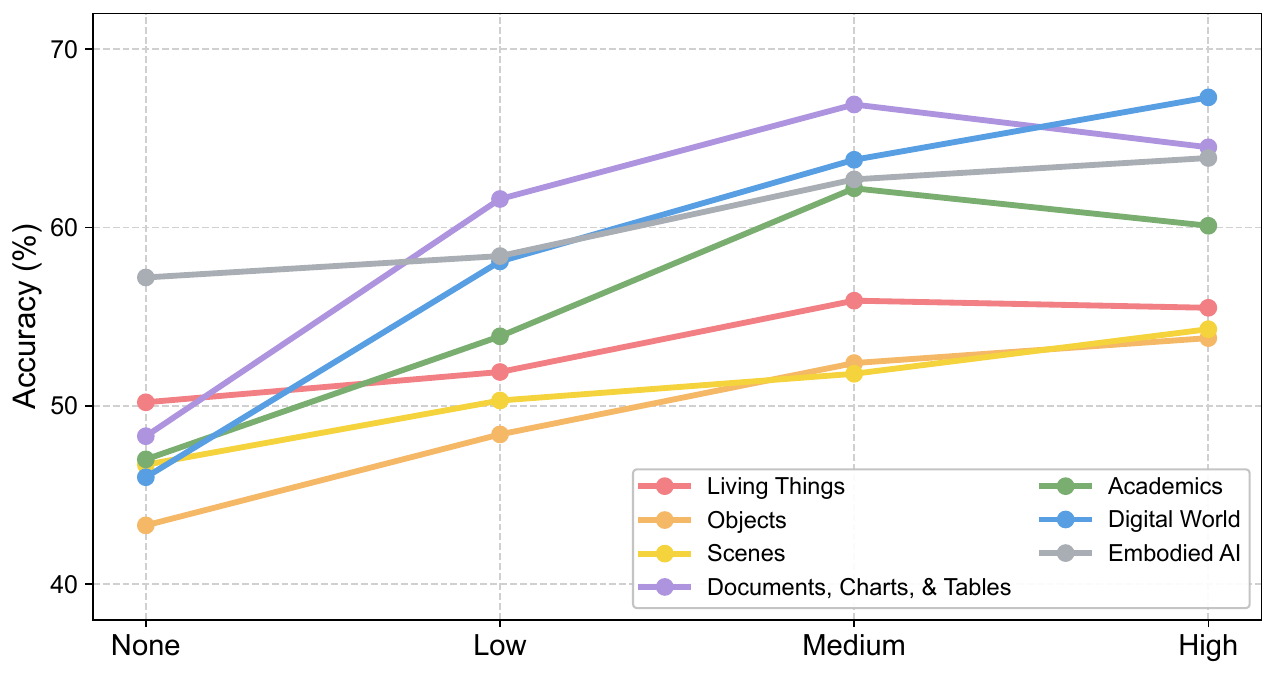}
        \caption{\captiontitle{Performance with longer reasoning} Increasing reasoning budget does not always increase performance on {\benchmarkname}, with even slight decrease on some domains.}
        \label{fig:reasoning}
    \end{minipage}\hfill
    \begin{minipage}[t]{0.49\linewidth}
        \centering
        \includegraphics[width=\linewidth]{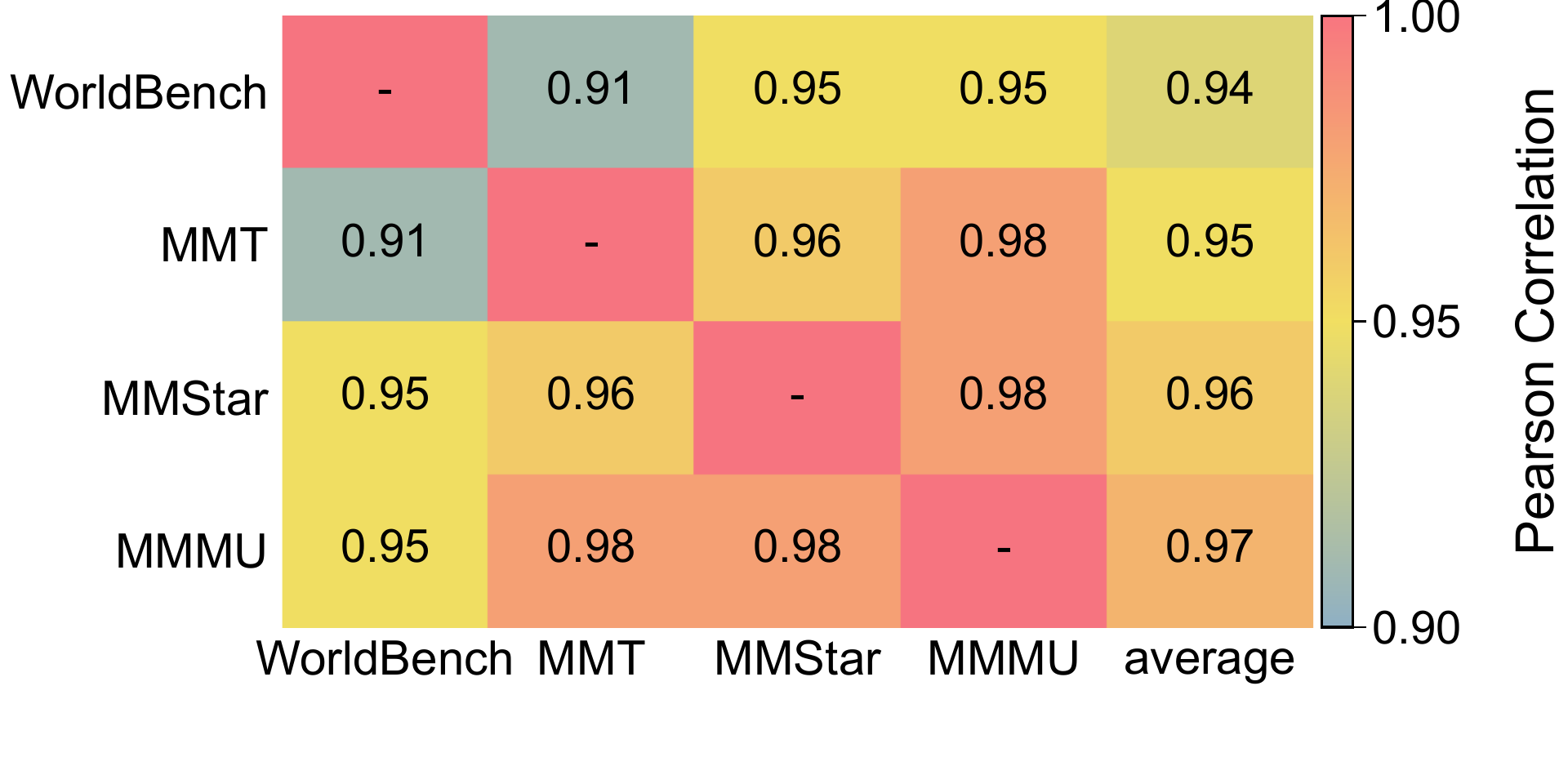}
        \caption{\captiontitle{{\benchmarkname} shows the lowest average correlation with other diverse benchmarks} We compute the pairwise correlation of model accuracies between each pair of benchmarks.}
        \label{fig:correlation}
    \end{minipage}
\end{figure}

\paragraph{Benefits of reasoning} Recent MLLMs~\citep{qwen3, vteam2025glm45vglm41vthinkingversatilemultimodal, kimiteam2025kimivltechnicalreport} have adopted Reinforcement Learning in post training~\citep{shao2024deepseekmathpushinglimitsmathematical} to enable the generation of intermediate \emph{Chain-of-Thought} reasoning steps~\citep{wei2023chainofthoughtpromptingelicitsreasoning} before producing final answers. To understand whether longer reasoning improves performance on {\benchmarkname}, we vary the reasoning budget of GPT-5.4 from none to high and measure accuracy across all domains. As shown in~\figref{fig:reasoning}, accuracy consistently improves from no reasoning to a low budget across all domains. Beyond a low budget, the trends diverge: {\digital}, {\objects}, {\scenes}, and {\agents} continue to climb monotonically through the high budget, while {\ocr}, {\academics}, and {\livingthings} saturate or regress at higher budgets. We find that generating more reasoning tokens can result in the model getting stuck in a reasoning loop.

\paragraph{Correlation with other benchmarks}
To understand how the capabilities assessed by {\benchmarkname} differ from other benchmarks, we compute the Pearson correlation between model accuracies on {\benchmarkname} and other benchmarks (using the local models in~\tabref{tab:evaluation}). We select three benchmarks from~\secref{sec:diversity}---\mmmu, \mmstar, and \mmt---that have different levels of visual diversity according to human evaluation (\figref{fig:human_evaluation}) and are not saturated while remaining manageable in size. In~\figref{fig:correlation}, {\benchmarkname} shows the lowest average correlation of 0.94 with others. This suggests that {\benchmarkname}'s emphasis on visual diversity offers a different dimension to evaluate MLLMs.

\section{Related Work}
\paragraph{Multimodal large language models}
The capabilities of multimodal Large Language Models (MLLMs) have drastically improved over the past years. Proprietary models~\citep{gpt54,gemini31,claude47opus,grok42} now achieve unprecedented performance on a wide range of visual understanding benchmarks. On the open-source side, progress has been equally rapid from early pioneers~\citep{dai2023instructblip,awadalla2023openflamingo,liu2023visual} to more recent models~\citep{qwen35, glm46, wang2025internvl3} that are increasingly competitive with frontier proprietary systems. Despite these advances, existing benchmarks~\citep{yue2024mmmu,tong2024eyeswideshutexploring} often evaluate models on image sets with limited visual diversity. Our work addresses this gap by evaluating MLLMs under a much broader and visually more diverse collection of images.

\paragraph{Multimodal benchmarks} Many multimodal benchmarks target specific image understanding capabilities of MLLMs, such as visual-centric perception tasks~\citep{paiss2023teaching,tong2024eyeswideshutexploring,fu2024blinkmultimodallargelanguage,tong2024cambrian1fullyopenvisioncentric}, OCR~\citep{wang2024charxiv,ma2024mmlongbenchdocbenchmarkinglongcontextdocument,fu2025ocrbenchv2improvedbenchmark}, object hallucination~\citep{li2023evaluatingobjecthallucinationlarge,guan2024hallusionbench}, GUI navigation~\citep{li2025screenspot,cheng2024seeclick,xie2025scaling,rawles2024androidworld}, and coding~\citep{si2025design2code,yang2025chartmimicevaluatinglmmscrossmodal}. Beyond such task-specific benchmarks, various benchmarks~\citep{LVLM-EHub,li2023seed,yu2024mmvet,liu2024mmbench,mmtbench,fu2023mme} aim to provide broader task coverage by comprehensively evaluating MLLMs across a diverse set of tasks. In contrast to all prior work, {\benchmarkname} departs from a task-centric design and is instead built around \emph{visual diversity}.
\section{Conclusion}

This work introduces {\benchmarkname}, a visually diverse reasoning benchmark to challenge MLLMs by seeing the whole world. We build a taxonomy with thousands of visual concepts, curate a diverse image set from it, and design questions through a structured trial-and-error process that targets failure modes of frontier MLLMs. Both quantitative and human evaluations show that {\benchmarkname} exhibits greater visual diversity than existing diverse benchmarks. Furthermore, our model evaluations reveal that {\benchmarkname} is very challenging for current models. We hope our work motivates building datasets that prioritize visual diversity.
{
    \small
    \bibliographystyle{ieeenat_fullname}
    \bibliography{main}
}

\clearpage
\appendix

\section{Question Annotation Interface}
\label{appendix:annotation_interface}
\begin{figure*}[h]
    \centering
    \includegraphics[width=\linewidth]{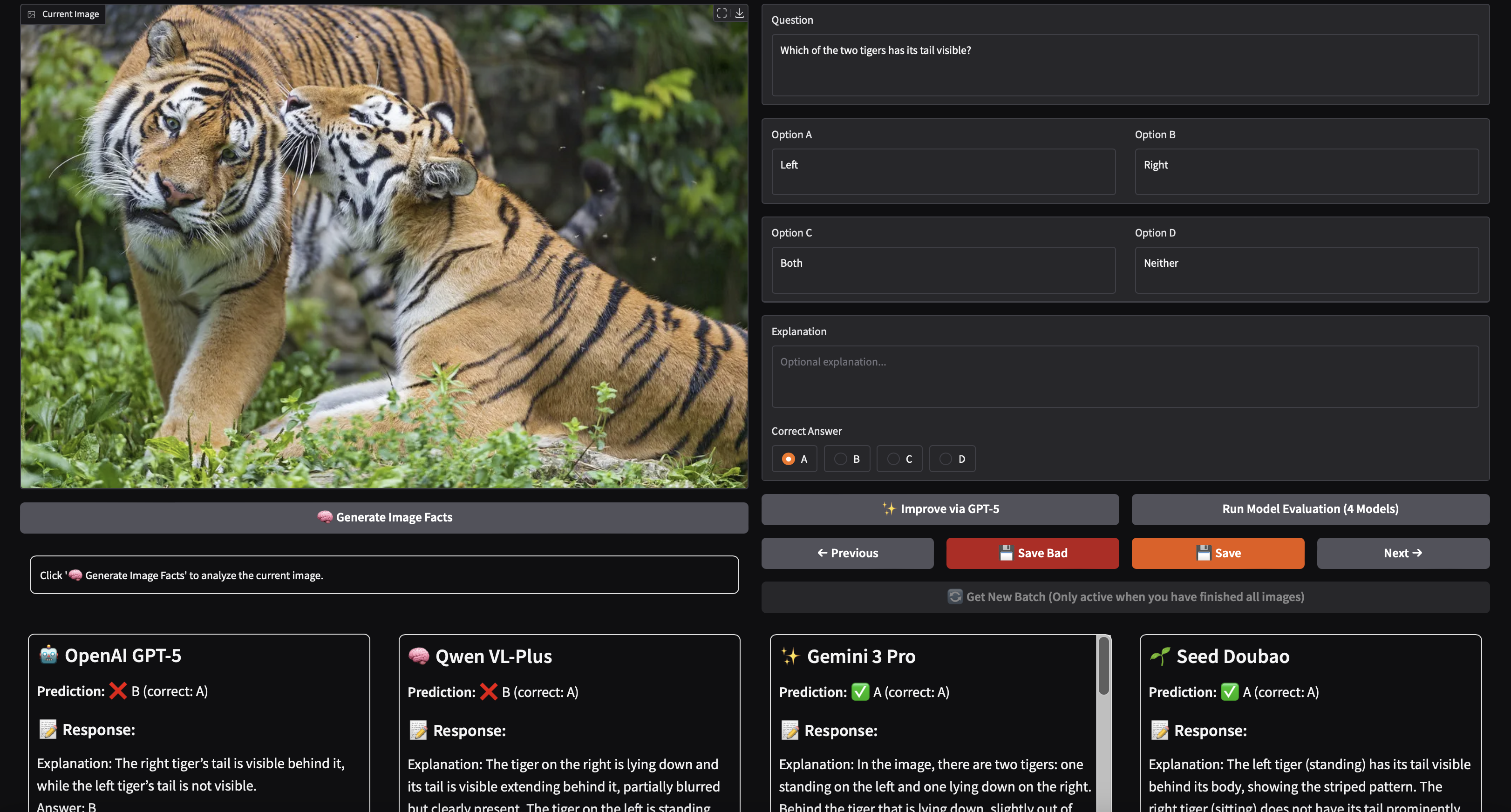}
    \caption{\captiontitle{Question annotation interface} Annotators use the interface to examine each image, propose a question, evaluate it against frontier models, and refine the question wording as well as explanation with an LLM.}
    \label{fig:annotation_interface}
\end{figure*}

\figref{fig:annotation_interface} illustrates our interface to design challenging questions. To help annotators in proposing questions, we provide a ``Generate Image Facts'' button that queries an LLM (\eg, GPT-5) to extract facts from an image. Annotators can also come up with questions in their own way. For each question, annotators are required to use the ``Run Model Evaluation (4 Models)'' button to check that at least one frontier model answers the question incorrectly. For questions that may confuse others or be mistakenly judged as incorrect, annotators need to provide a brief explanation clarifying the correct answer. The interface also includes an ``Improve via GPT-5'' button that automatically refines the wording of the question, answer choices, and explanation using an LLM (\eg, GPT-5).

\section{Additional Results}
\label{appendix:additional_results}
In~\secref{sec:model_evaluation}, we evaluate the visual diversity of images in existing multimodal benchmarks. We now extend this to several widely used pre-training datasets, including YFCC~\citep{Thomee2016}, CC~\citep{changpinyo2021cc12m}, DataComp~\citep{gadre2023datacomp}, WIT~\citep{srinivasan2021wit}, LAION~\citep{schuhmann2022laionb}, and ImageNet~\citep{Deng2009}, to examine how different vision encoders assess their visual diversity. For each dataset, we randomly sample 1,000 images and compute visual diversity using effective rank and participation ratio across different pre-trained vision encoders. \tabref{tab:embedding_pretrain_dataset} reports the results.

\begin{table*}[h]
    \centering
    \tablestyle{4pt}{1.0}
    \begin{tabular}{lcccccc}
    \multirow{2}{*}{Dataset} & \multicolumn{2}{c}{SigLIP 2} & \multicolumn{2}{c}{Perception Encoder} & \multicolumn{2}{c}{DINO v3} \\
      & Effective Rank & Participation Ratio & Effective Rank & Participation Ratio & Effective Rank & Participation Ratio \\
    \shline
    YFCC & 17.9 & 103.9 & 27.7 & 133.7 & 572.1 & 540.5 \\
    CC & 42.1 & 146.3 & \textbf{50.9} & 189.3 & 542.6 & 525.1 \\
    DataComp & \textbf{45.9} & \textbf{154.1} & 49.7 & \textbf{193.8} & 516.8 & 511.9 \\
    WIT & 19.8 & 78.0 & 29.3 & 108.2 & 453.2 & 293.1 \\
    LAION & 48.5 & 146.8 & 49.3 & 170.7 & 452.1 & 393.1 \\
    ImageNet & 24.9 & 107.1 & 30.2 & 120.3 & \textbf{613.8} & \textbf{610.4} \\
    \end{tabular}
    \caption{\captiontitle{Quantitative evaluation on pre-training datasets} We observe that vision encoders yield higher diversity scores on datasets that are similar to their pre-training distribution.}
    \label{tab:embedding_pretrain_dataset}
\end{table*}
We observe that different vision encoders rank the same datasets differently. SigLIP2 considers DataComp the most diverse, whereas DINO~v3 rates ImageNet highest. This is likely influenced by differences in pre-training data distributions of these encoders. SigLIP2 is trained on WebLI, a proprietary dataset sourced from the Internet. This may lead its learned representations to be more aligned with other web-scale datasets, such as DataComp from the web archive data Common Crawl. In contrast, DINO~v3 is trained on a mixture of Instagram images and ImageNet, with ImageNet comprising 1/10 of its pre-training data. As a result, this encoder assigns higher diversity scores to ImageNet. Overall, these results suggest that datasets more similar to an encoder's pre-training distribution tend to receive higher visual diversity scores under that encoder. More broadly, visual diversity is difficult to capture with any single metric or representation; our human evaluation provides a complementary signal, but future work should scale it to a larger and more demographically diverse set of judgments. {\benchmarkname} can help identify weaknesses in MLLMs across visually diverse settings, but because it is curated from web search results and existing datasets, it may reflect biases in those sources and should not be treated as sufficient evidence of safe or fair real-world deployment.

\section{User Study Interface for Human Evaluation}
\begin{figure*}[h]
    \centering
    \includegraphics[width=\linewidth]{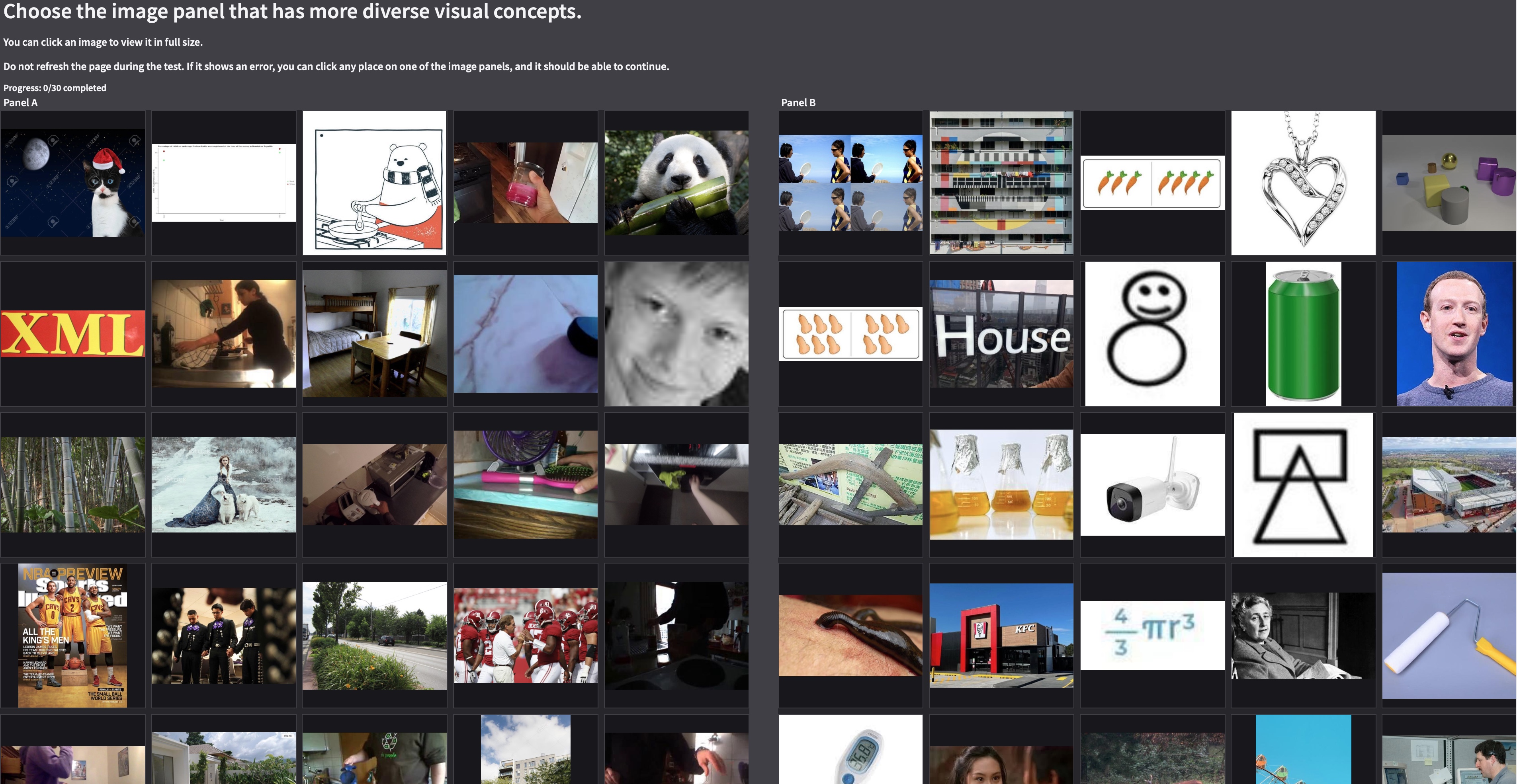}
    \caption{\captiontitle{User study interface for human evaluation} We display two image panels side by side and ask users to choose the one that is visually more diverse.}
    \label{fig:human_evaluation_interface}
\end{figure*}

\figref{fig:human_evaluation_interface} illustrates our interface to conduct human evaluation of visual diversity across benchmarks. For each comparison, we randomly sample 100 images from two benchmarks and display them side-by-side in a 20x5 image grid. Users can click any image to view it in full resolution. They are then asked to carefully examine all images on each side and choose the panel that is visually more diverse. 

\section{Implementation Details}
\label{appendix:implementation}
\subsection{Benchmarks Used in Diversity Evaluation}
\label{appendix:implementation_benchmark}
\paragraph{Removing similar images}
Some benchmarks we evaluate in~\secref{sec:diversity} contain identical images reused across different questions or multiple similar images within a single question. Using all images in such cases would reduce their visual diversity and place them at a disadvantage relative to {\benchmarkname}.

Instead, we randomly sample only one image per question for each benchmark and then remove any \emph{exact duplicate} within this set. For \mega, we do not use any images from their in-context examples; we only sample from the images that models are actually evaluated on.

\paragraph{Benchmarks}
We provide details on benchmarks we use in \secref{sec:diversity}. We use the validation split of \vqa~\citep{goyal2017making}, the full set of \mme~\citep{fu2023mme}, the full set of \mmstar~\citep{chen2024we}, the test split of \mmbench~\citep{yu2023mm}, the test split of \mmmu~\citep{yue2024mmmu}, the full set of \seedbench~\citep{li2023seedbench2benchmarkingmultimodallarge}, the test split of \mmt~\citep{mmtbench}, and the full set of \mega~\citep{chen2025megabench}. 

During {\benchmarkname} curation, we filter web-search results for Creative Commons-licensed images and manually review images and questions to remove unsafe, private, or otherwise inappropriate content before release.

\subsection{Model Performance}
\label{appendix:implementation_model}
\paragraph{Evaluation details} We report evaluation settings for proprietary models and open-source models. For proprietary models, we evaluate GPT-5.4-Thinking~\citep{gpt54}, Gemini-3.1-Pro~\citep{gemini31}, Gemini-3-Flash~\citep{gemini3}, Claude-Opus-4.7~\citep{claude47opus}, Grok-4.2~\citep{grok42}, and Qwen3.5-VL-Plus-Thinking/Instruct~\citep{qwen35}. We use their default sampling settings to generate responses, except where a model exposes a reasoning-effort setting reported in the model name.

For open-source models, we run offline inference locally with vLLM~\citep{kwon2023efficient} for efficiency, except for Kimi-K2.5, which we evaluate through its API due to its size. Each local model evaluation uses approximately 4--8 NVIDIA A100 GPUs and takes roughly one hour, depending on model size. We use each local model's default or officially recommended generation parameters (\eg, temperature, top-p, top-k). For reproducibility, we report the parameters used for local inference in~\tabref{tab:sampling}.

\begin{table}[h]
    \centering
    \tablestyle{5pt}{1.0}
    \begin{tabular}{l|ccc}
        models & temperature & top-p & top-k\\
        \Xhline{0.7pt}
        Qwen3.5-VL-35B-A3B~\citep{qwen35} & 1.0 & 0.95 & 20 \\
        Qwen3.5-VL-27B~\citep{qwen35} & 1.0 & 0.95 & 20 \\
        GLM-4.6V~\citep{glm46} & 0.8 & 0.6 & 2 \\
        Gemma-4-31B~\citep{gemma4} & 1.0 & 0.95 & 64 \\
        Gemma-4-E4B~\citep{gemma4} & 1.0 & 0.95 & 64 \\
        InternVL3.5~\citep{wang2025internvl3} & 0.6 & 1.0 & 50 \\
    \end{tabular}
    \caption{\textbf{Sampling parameters for local open-source model inference.} Kimi-K2.5 is excluded because it is evaluated through its API.}
    \label{tab:sampling}
\end{table}

\begin{figure}[h]
    \centering
    \includegraphics[width=.5\linewidth]{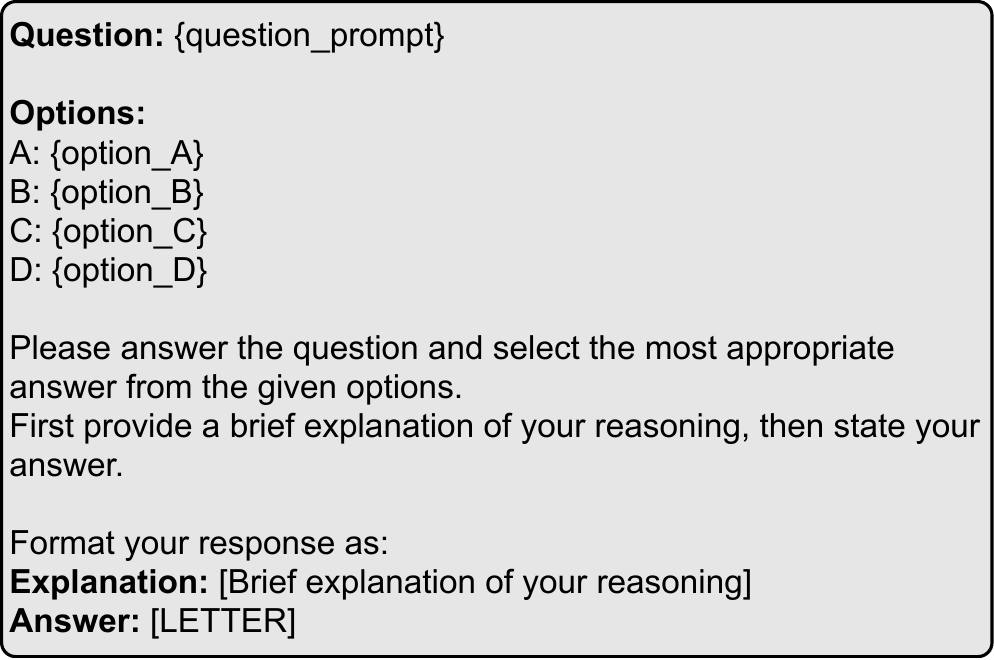}
    \caption{\captiontitle{Prompt to format question and answer options} We encourage the model to produce a short explanation before giving the final answer. This yields slightly higher accuracy than asking for the answer before an explanation.}
    \label{fig:prompt_template}
\end{figure}

\paragraph{Prompt template and answer parsing} \figref{fig:prompt_template} illustrates our prompt to format each question prompt and its answer choices. The prompt asks the model first to provide a brief explanation \emph{before} outputting the final answer. We find that this ordering slightly improves accuracy compared to asking for the final answer before an explanation. In the latter case, models sometimes change to a different answer during the explanation. 

To robustly extract the model's chosen answer option, we apply three regular expression rules one by one. We first normalize the response to uppercase and search for an explicit ``Answer: X'' pattern, where X $\in$ \{\text{A}, \text{B}, \text{C}, \text{D}\}. If this fails, we fall back to parsing a single option letter at the beginning of the response, and finally to detecting a standalone option letter, anywhere in the output. If none of them match, we consider the response incorrect.
Our parsing method achieves an average success rate of >99\% across all questions in {\benchmarkname}. As a result, we do not rely on more advanced approaches (\eg, LLM-as-a-judge).

\section{Model Responses}

\begin{figure*}[p]
    \centering
    \includegraphics[width=\linewidth]{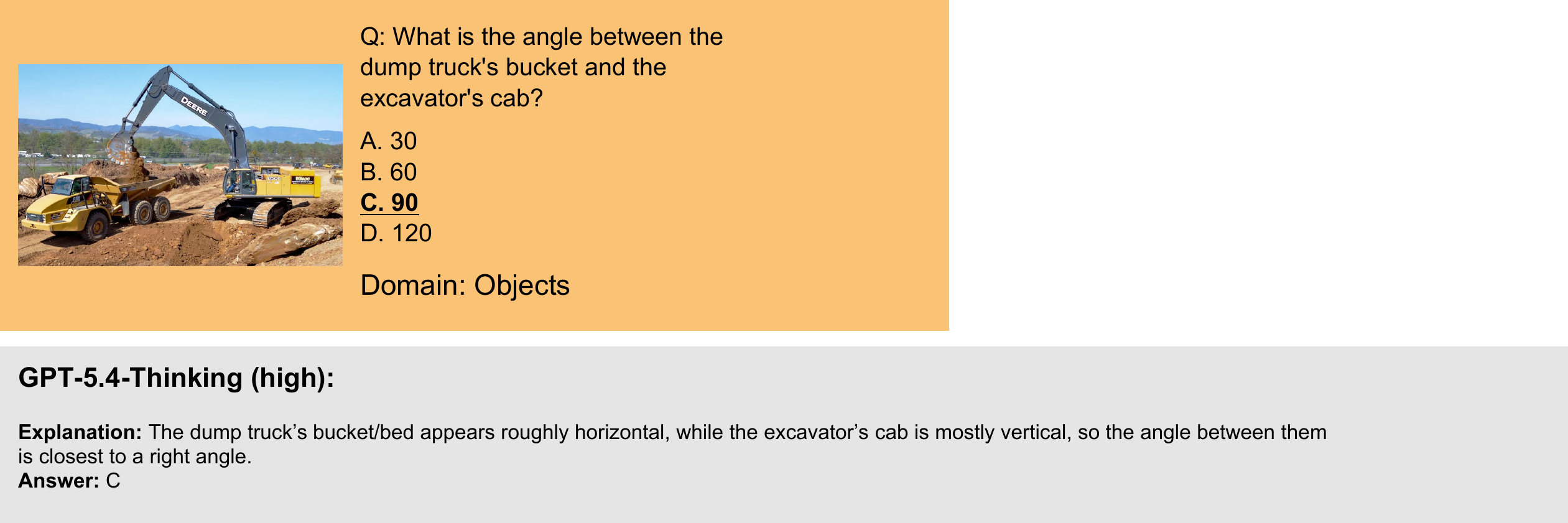}
    \caption{Example response from GPT-5.4-Thinking (high)~\citep{gpt54} on an {\objects} question.}
    \label{fig:gpt54_thinking_high_response}
\end{figure*}

\begin{figure*}[p]
    \centering
    \includegraphics[width=\linewidth]{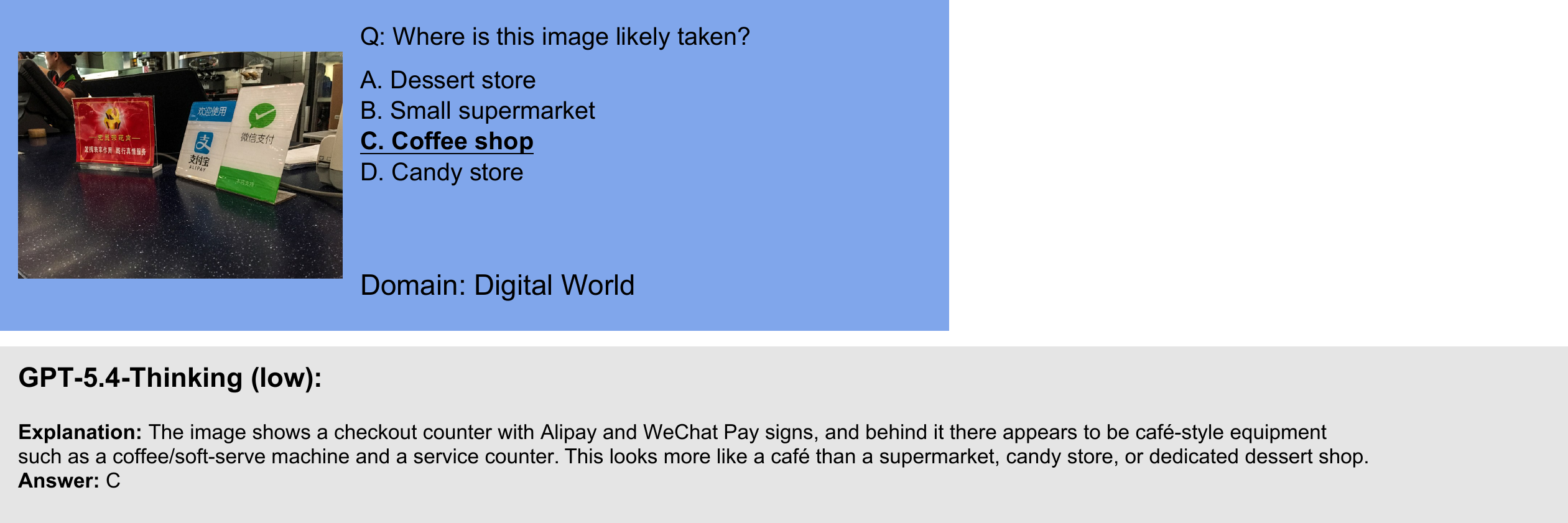}
    \caption{Example response from GPT-5.4-Thinking (low)~\citep{gpt54} on a {\digital} question.}
    \label{fig:gpt54_thinking_low_response}
\end{figure*}

\begin{figure*}[p]
    \centering
    \includegraphics[width=\linewidth]{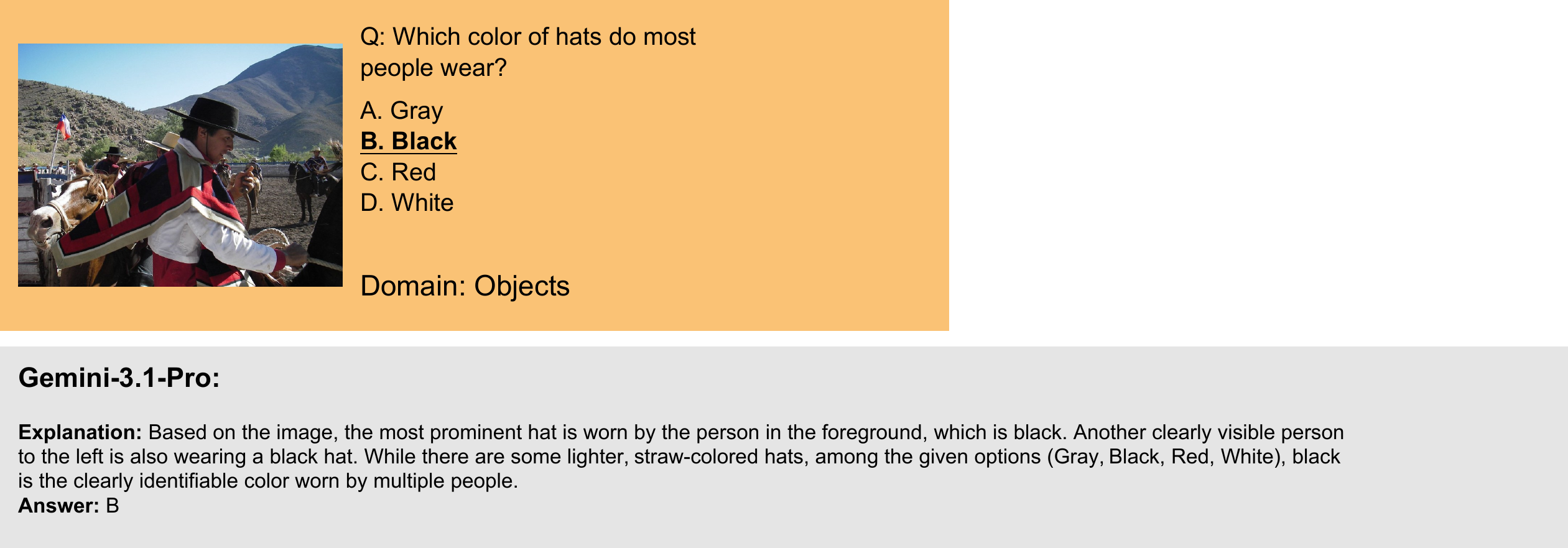}
    \caption{Example response from Gemini-3.1-Pro~\citep{gemini31} on an {\objects} question.}
    \label{fig:gemini_31_pro_response}
\end{figure*}

\begin{figure*}[p]
    \centering
    \includegraphics[width=\linewidth]{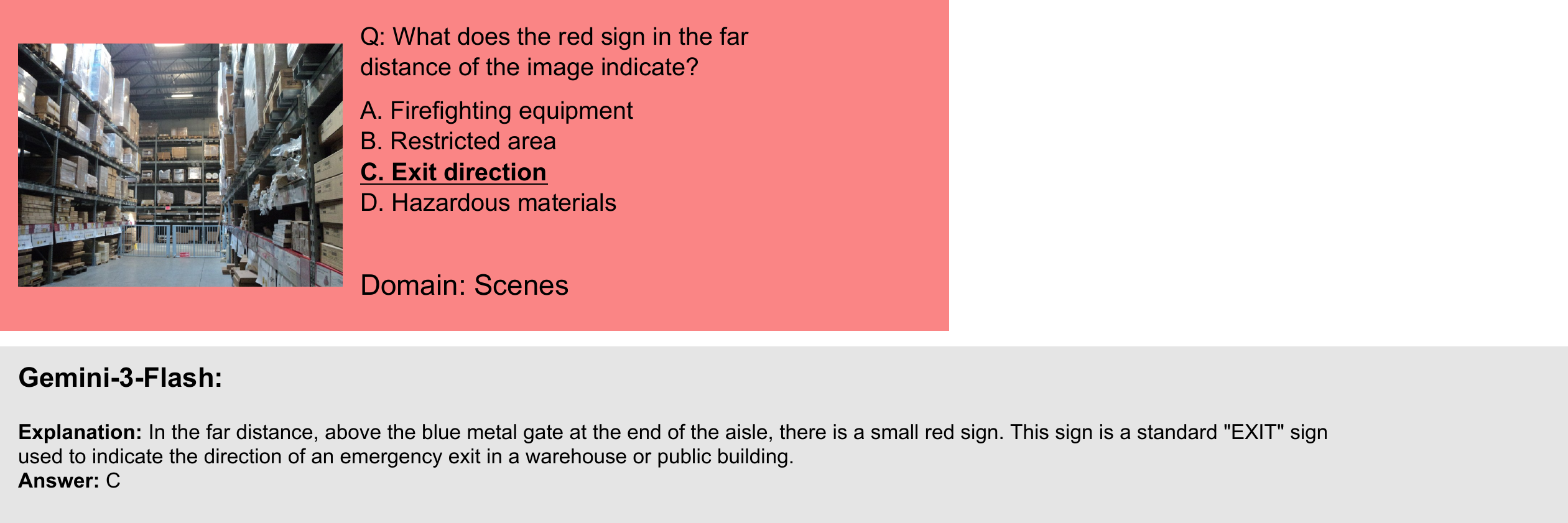}
    \caption{Example response from Gemini-3-Flash~\citep{gemini3} on a {\scenes} question.}
    \label{fig:gemini_3_flash_response}
\end{figure*}

\begin{figure*}[p]
    \centering
    \includegraphics[width=\linewidth]{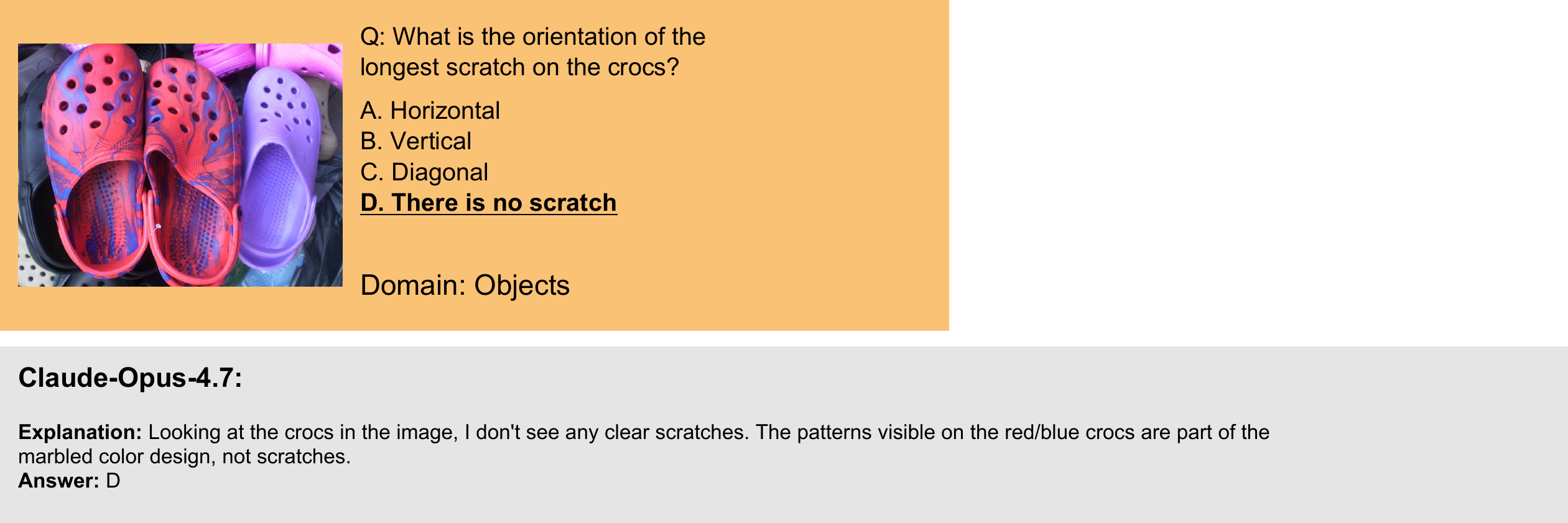}
    \caption{Example response from Claude-Opus-4.7~\citep{claude47opus} on an {\objects} question.}
    \label{fig:claude_opus_47_response}
\end{figure*}

\begin{figure*}[p]
    \centering
    \includegraphics[width=\linewidth]{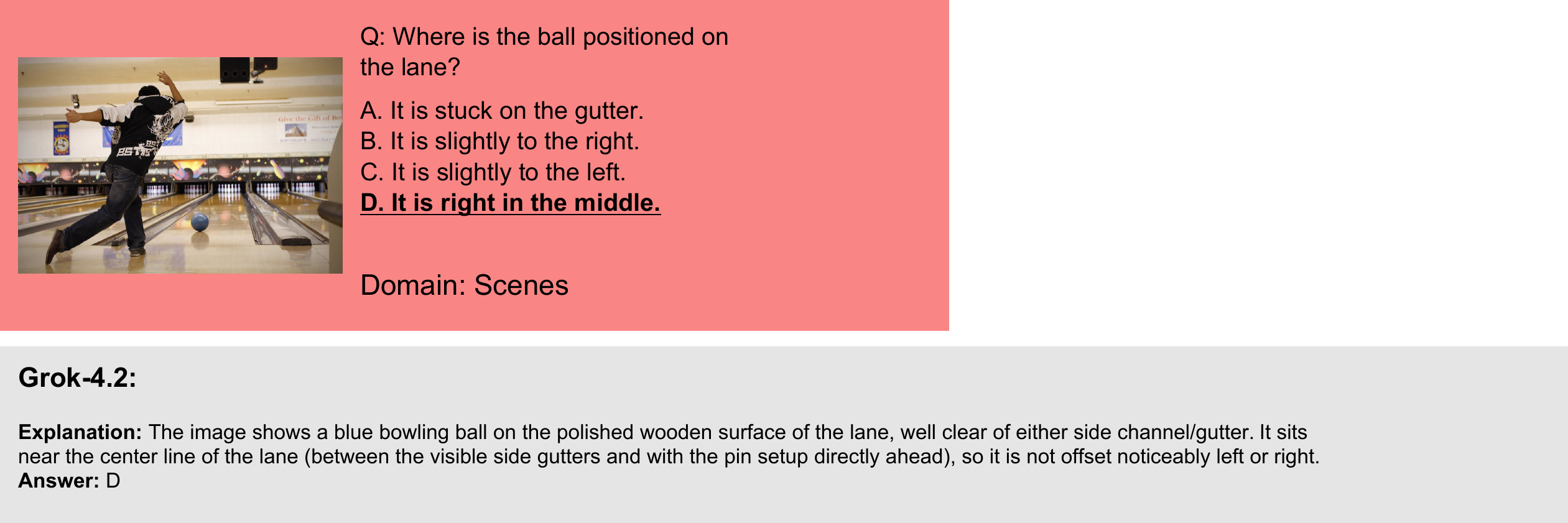}
    \caption{Example response from Grok-4.2~\citep{grok42} on a {\scenes} question.}
    \label{fig:grok_42_response}
\end{figure*}

\begin{figure*}[p]
    \centering
    \includegraphics[width=\linewidth]{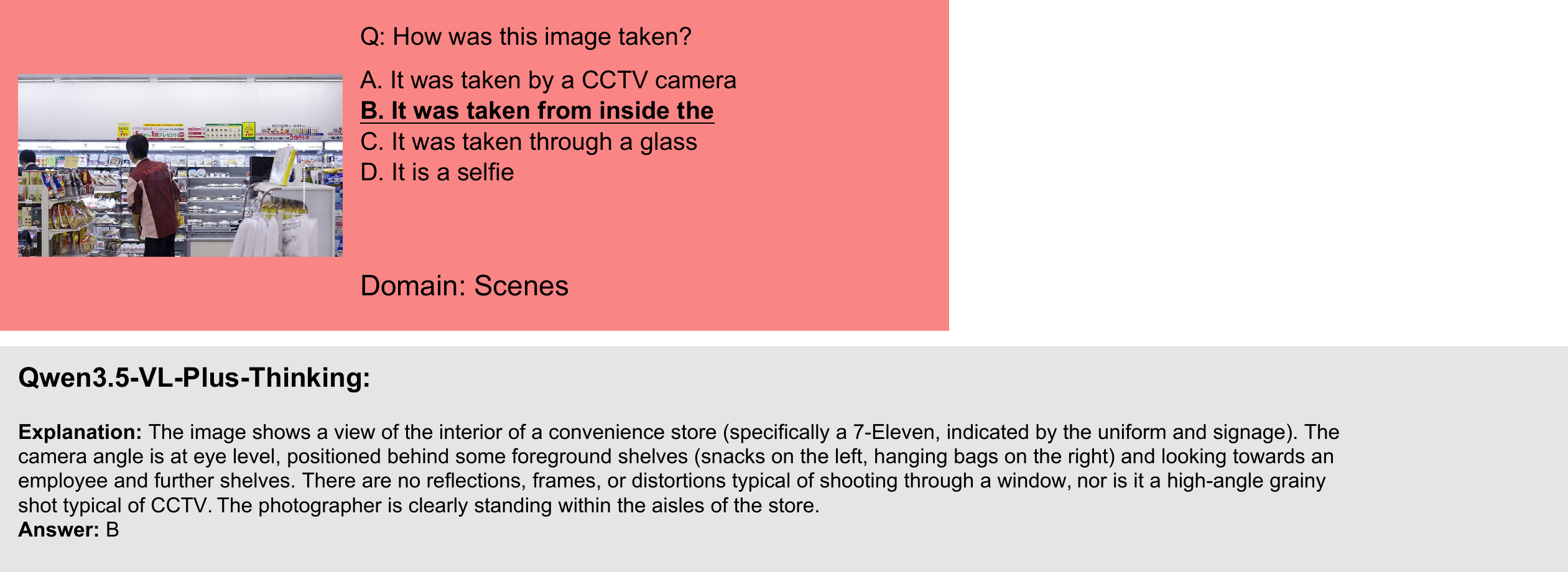}
    \caption{Example response from Qwen3.5-VL-Plus-Thinking~\citep{qwen35} on a {\scenes} question.}
    \label{fig:qwen35_plus_thinking_response}
\end{figure*}

\begin{figure*}[p]
    \centering
    \includegraphics[width=\linewidth]{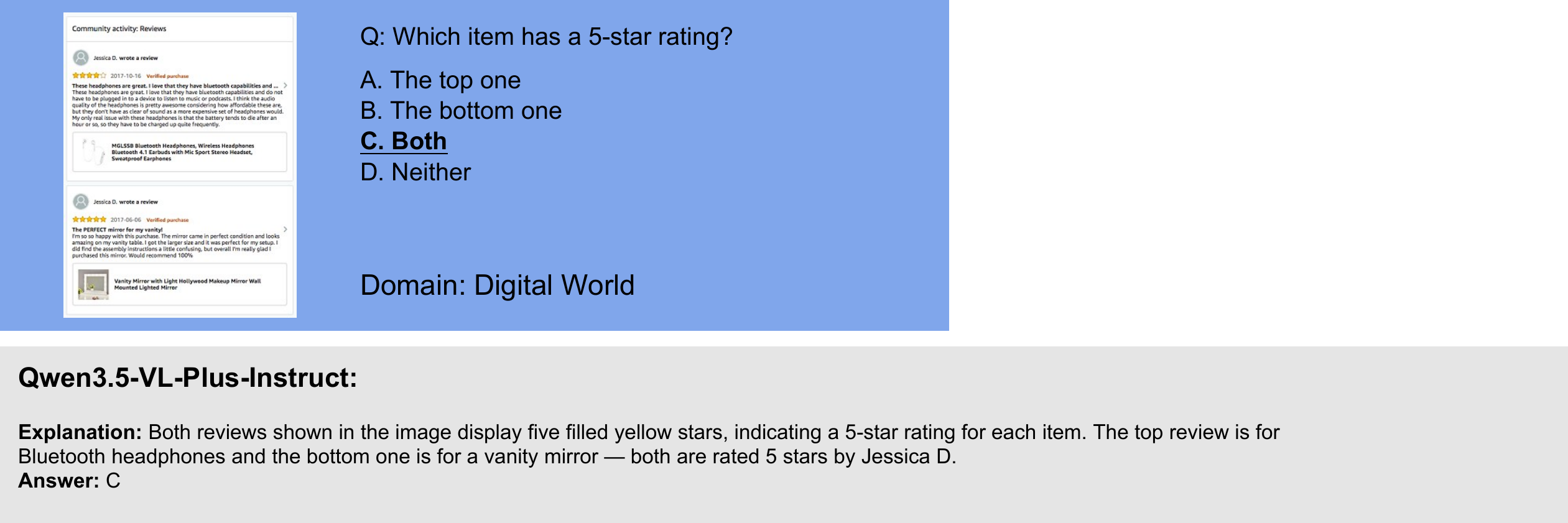}
    \caption{Example response from Qwen3.5-VL-Plus-Instruct~\citep{qwen35} on a {\digital} question.}
    \label{fig:qwen35_plus_instruct_response}
\end{figure*}

\begin{figure*}[p]
    \centering
    \includegraphics[width=\linewidth]{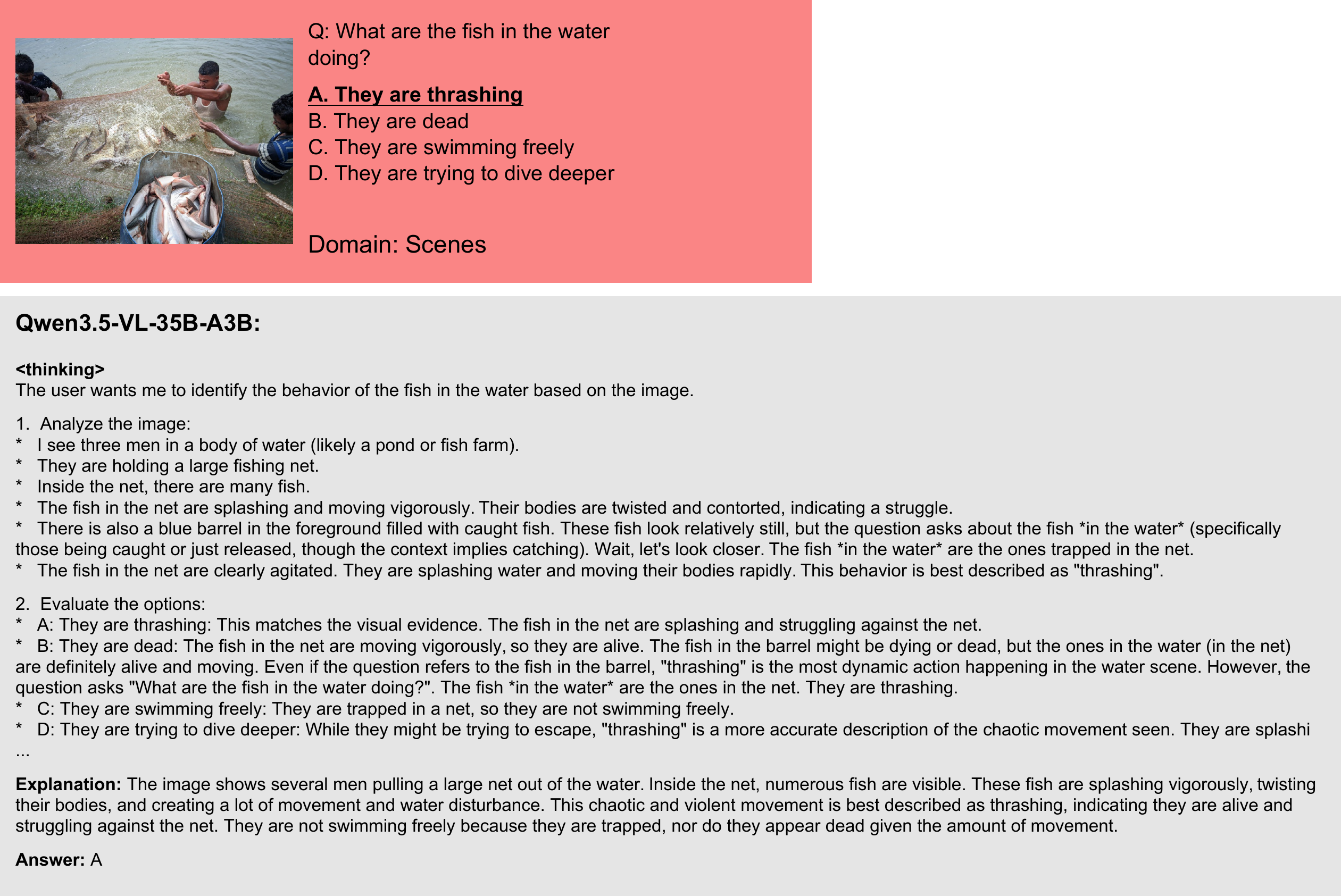}
    \caption{Example response from Qwen3.5-VL-35B-A3B~\citep{qwen35} on a {\scenes} question.}
    \label{fig:qwen35_vl_35b_a3b_response}
\end{figure*}

\begin{figure*}[p]
    \centering
    \includegraphics[width=\linewidth]{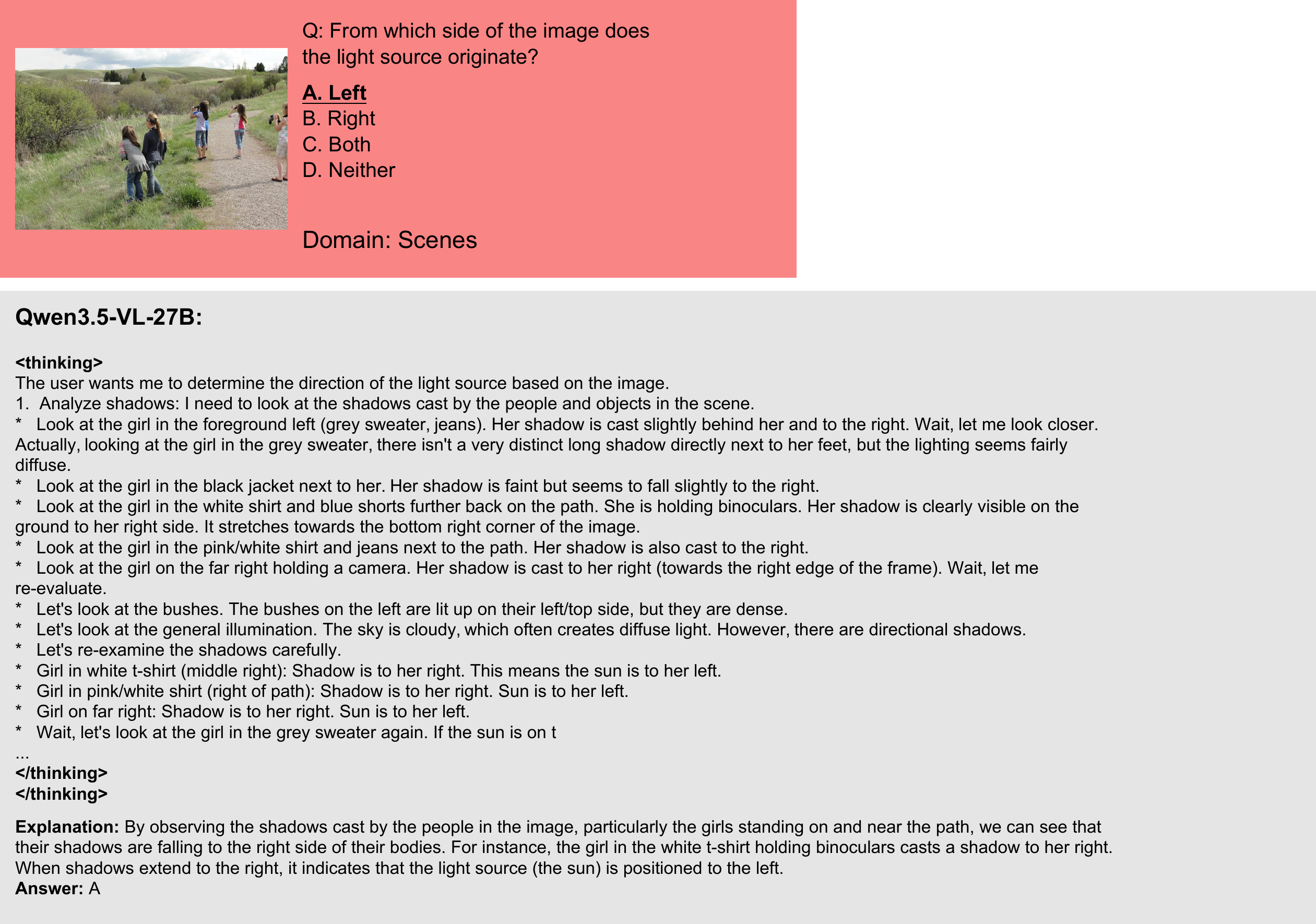}
    \caption{Example response from Qwen3.5-VL-27B~\citep{qwen35} on a {\scenes} question.}
    \label{fig:qwen35_vl_27b_response}
\end{figure*}

\begin{figure*}[p]
    \centering
    \includegraphics[width=\linewidth]{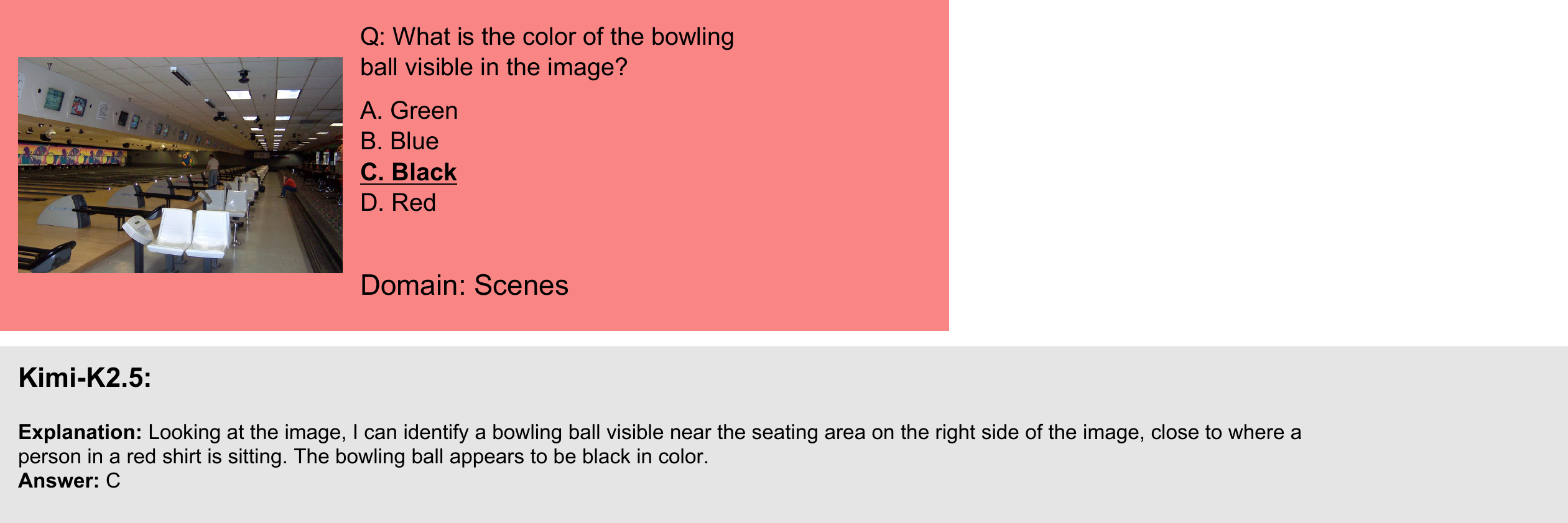}
    \caption{Example response from Kimi-K2.5~\citep{team2026kimi} on a {\scenes} question.}
    \label{fig:kimi_k25_response}
\end{figure*}

\begin{figure*}[p]
    \centering
    \includegraphics[width=\linewidth]{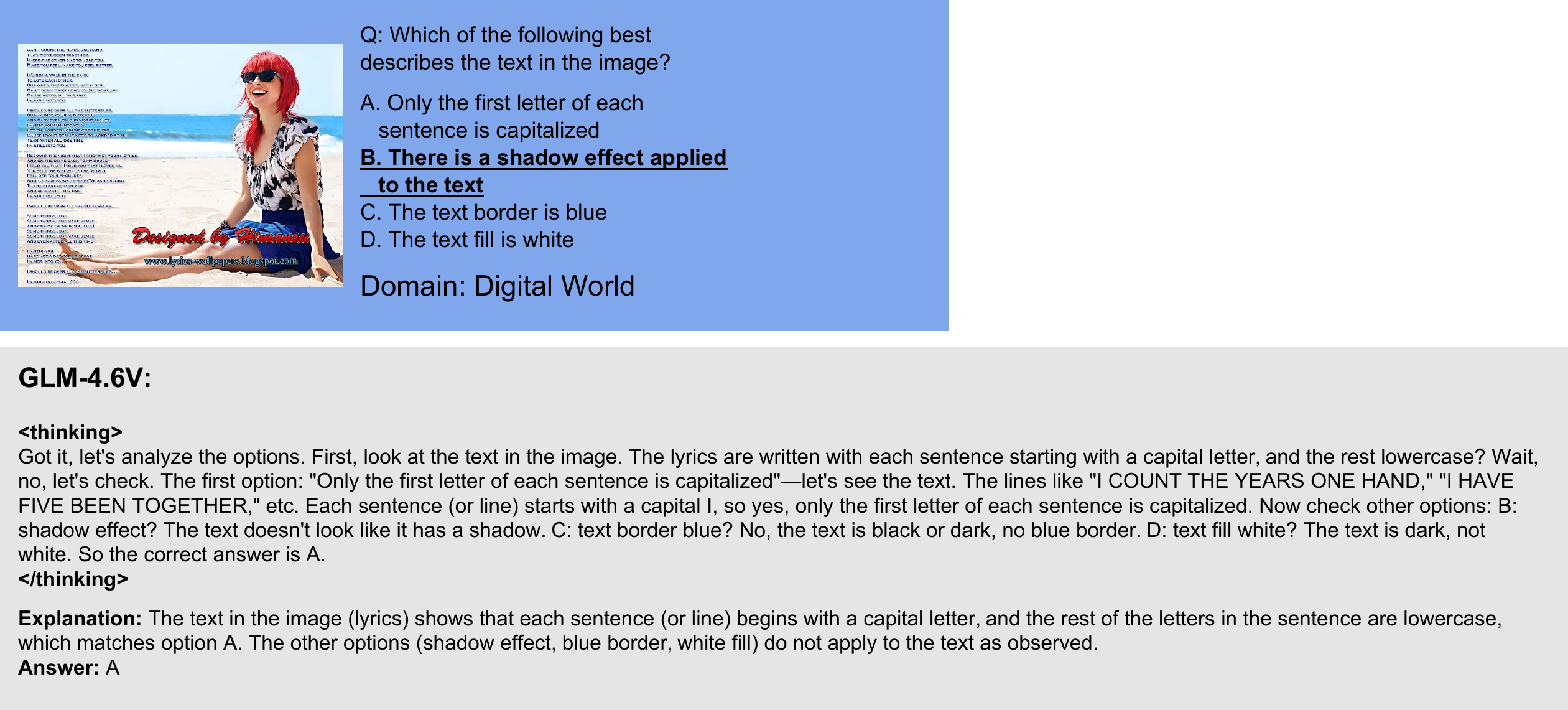}
    \caption{Example response from GLM-4.6V~\citep{glm46} on a {\digital} question.}
    \label{fig:glm_46v_response}
\end{figure*}

\begin{figure*}[p]
    \centering
    \includegraphics[width=\linewidth]{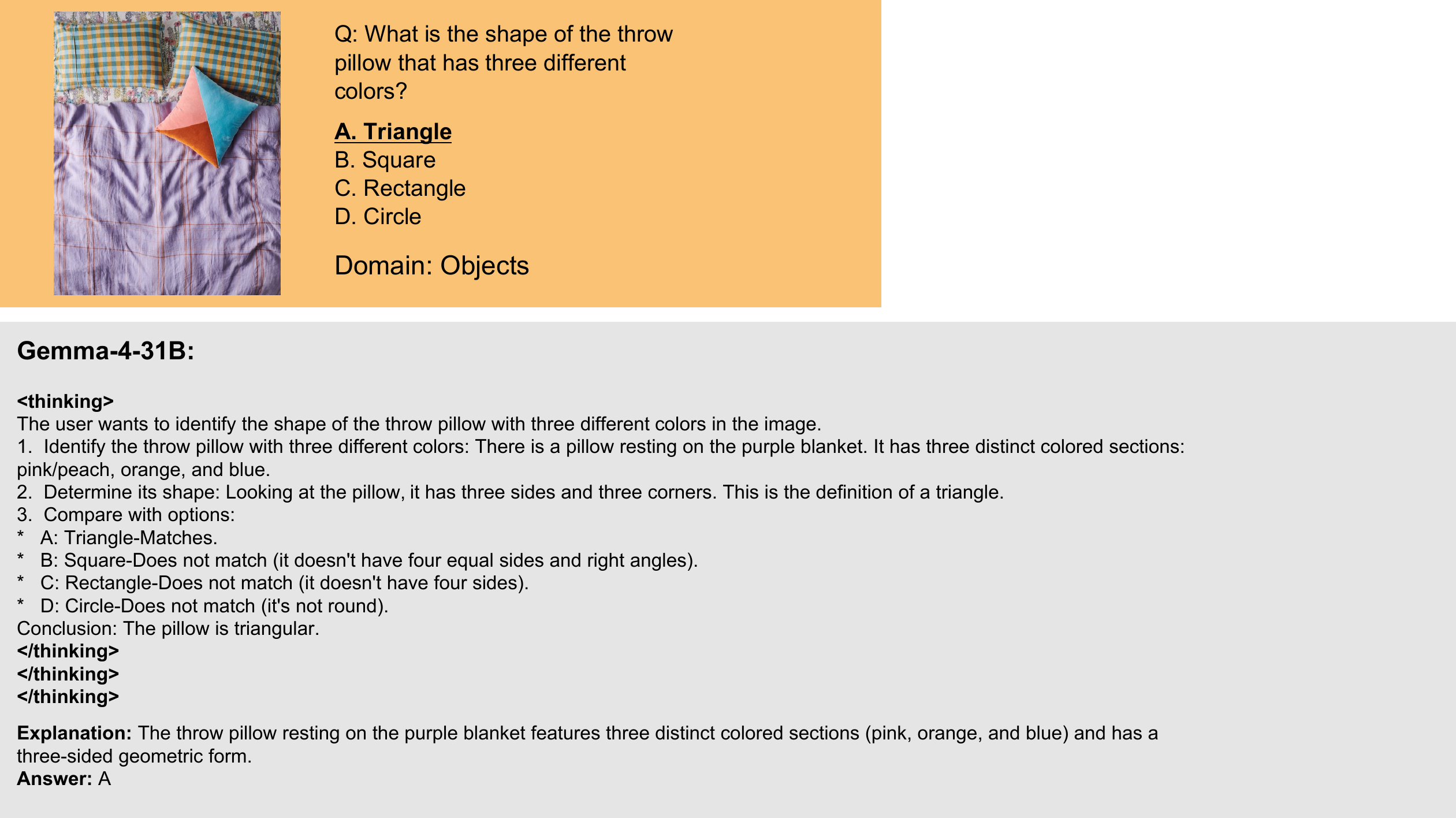}
    \caption{Example response from Gemma-4-31B~\citep{gemma4} on an {\objects} question.}
    \label{fig:gemma4_31b_response}
\end{figure*}

\begin{figure*}[p]
    \centering
    \includegraphics[width=\linewidth]{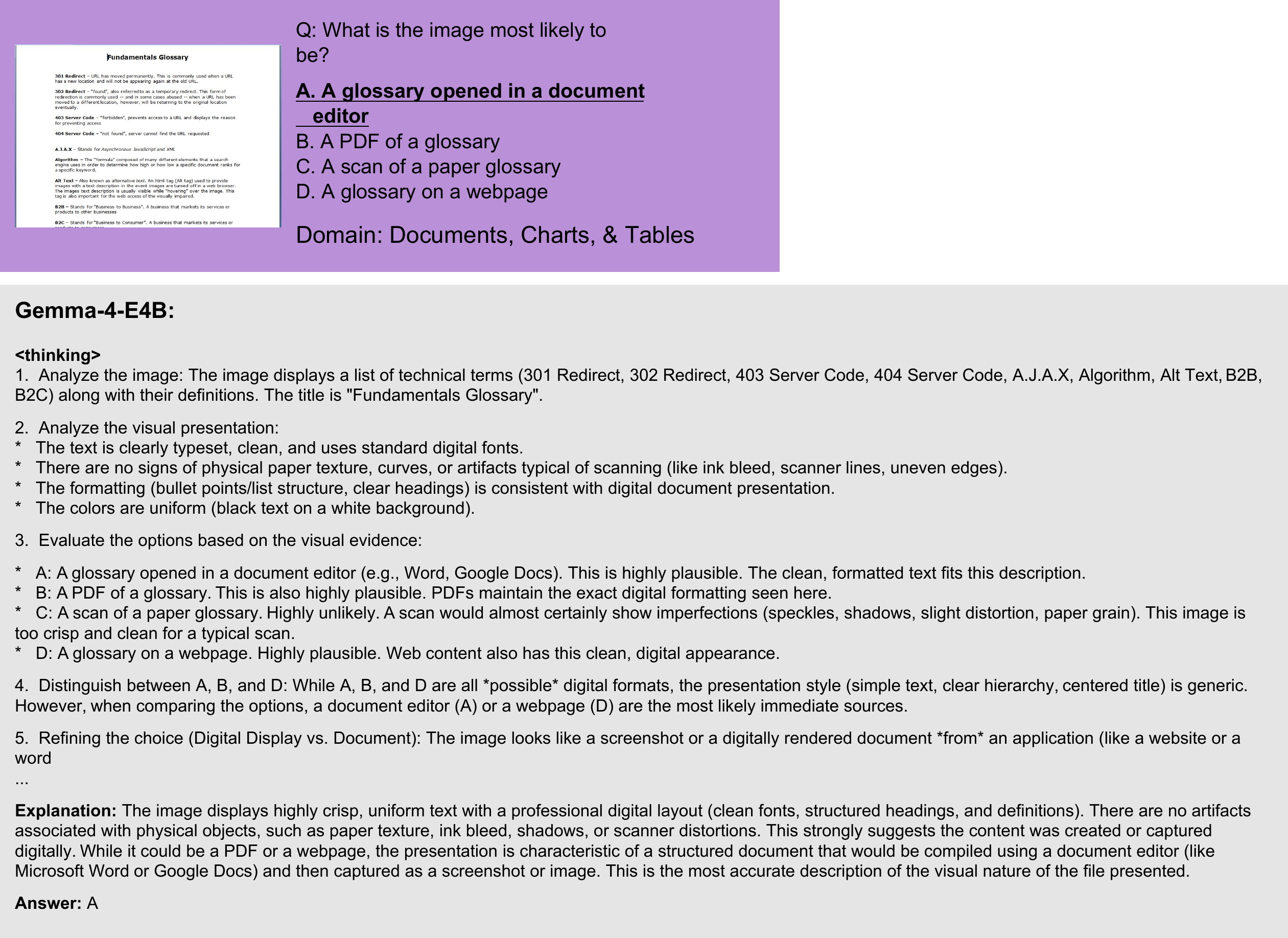}
    \caption{Example response from Gemma-4-E4B~\citep{gemma4} on a {\ocr} question.}
    \label{fig:gemma4_e4b_response}
\end{figure*}

\begin{figure*}[p]
    \centering
    \includegraphics[width=\linewidth]{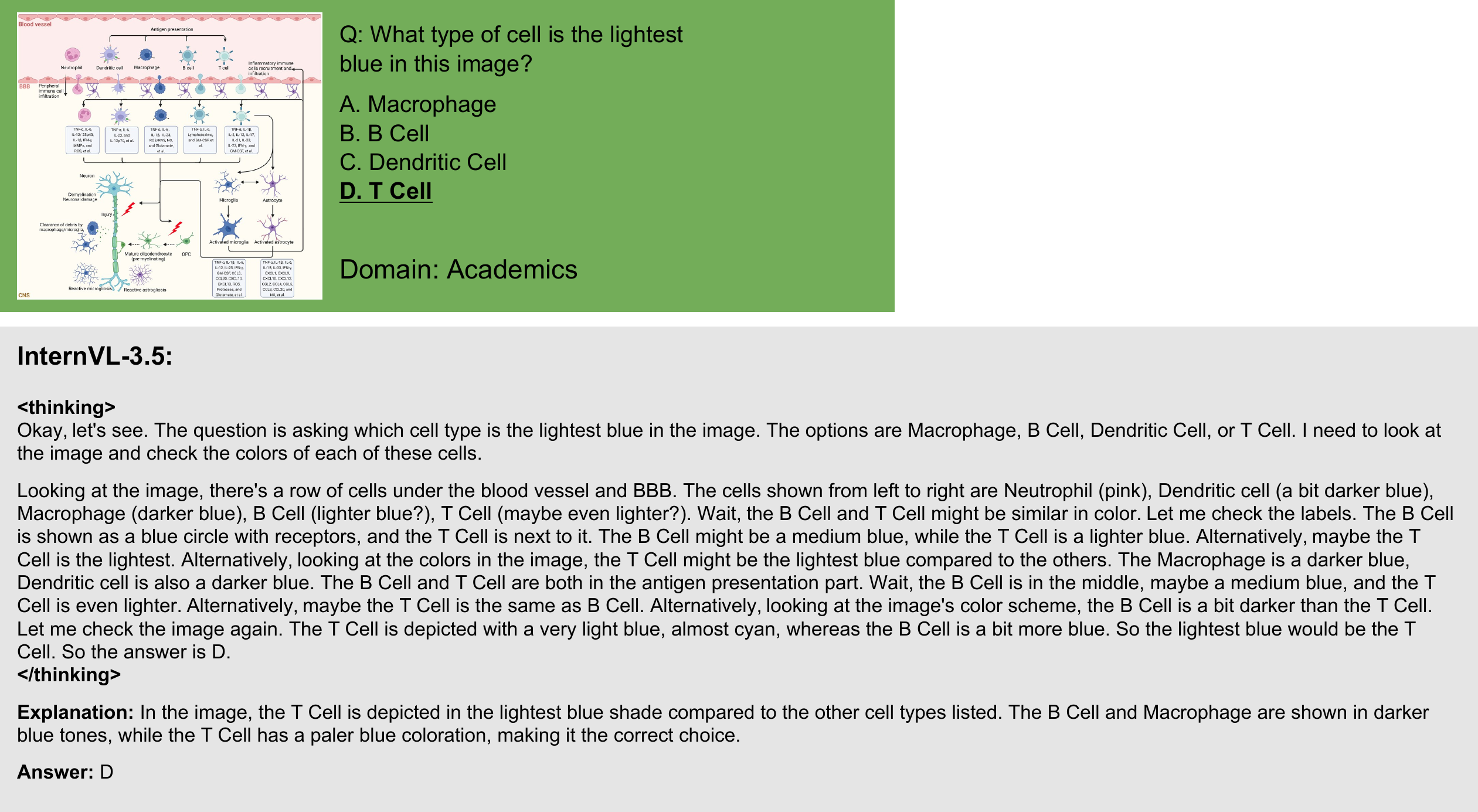}
    \caption{Example response from InternVL-3.5~\citep{wang2025internvl3} on an {\academics} question.}
    \label{fig:internvl35_response}
\end{figure*}

We present example responses to {\benchmarkname} questions from all 15 models evaluated in Section~\ref{sec:model_performance}. Figures~\ref{fig:gpt54_thinking_high_response}, \ref{fig:gpt54_thinking_low_response}, \ref{fig:gemini_31_pro_response}, \ref{fig:gemini_3_flash_response}, \ref{fig:claude_opus_47_response}, \ref{fig:grok_42_response}, \ref{fig:qwen35_plus_thinking_response}, \ref{fig:qwen35_plus_instruct_response}, \ref{fig:qwen35_vl_35b_a3b_response}, \ref{fig:qwen35_vl_27b_response}, \ref{fig:kimi_k25_response}, \ref{fig:glm_46v_response}, \ref{fig:gemma4_31b_response}, \ref{fig:gemma4_e4b_response}, and \ref{fig:internvl35_response} each illustrate one representative output from the corresponding model. We include exposed reasoning traces when they are available; some thinking models do not return their private reasoning traces through the API.


\end{document}